\PassOptionsToPackage{dvipsnames}{xcolor}
\documentclass[10pt]{article}
\newif\ifarxiv
\arxivtrue
\ifarxiv
\usepackage[preprint]{tmlr}
\else
\usepackage{tmlr}
\fi
\usepackage[hyphens]{url}
\usepackage{graphicx}
\usepackage{caption}

\usepackage{amsmath}
\usepackage{amssymb}
\usepackage{mathtools}
\usepackage{mathrsfs}
\usepackage{amsthm}
\usepackage{pifont}
\usepackage{xcolor}
\usepackage{enumitem}
\usepackage{booktabs}
\usepackage{adjustbox}

\usepackage{tikz}
\usetikzlibrary{matrix,fit,backgrounds,calc}
\definecolor{ao}{rgb}{0.0, 0.3, 1.0}
\definecolor{bittersweet}{rgb}{1.0, 0.44, 0.37}

\newcommand{\fS}{\mathcal{S}}
\newcommand{\fA}{\mathcal{A}}

\newcommand{\E}{\mathbb{E}}

\newenvironment{tightitems}{
  \vspace{-3pt}
  \begin{itemize}[leftmargin=10pt, labelsep=1pt, itemsep=-1pt]
}{
  \end{itemize}
  \vspace{-3pt}
}

\newtheorem{proposition}{Proposition}

\usepackage{hyperref}

\title{Algorithm Distillation for Safe In-Context Reinforcement Learning}
\ifarxiv
\author{\name Minjae Kwon$^{*}$ \email mjkwon@virginia.edu \\
      \addr University of Virginia
      \AND
      \name Amir Moeini$^{*}$ \email amoeini@virginia.edu \\
      \addr University of Virginia
      \AND
      \name Shangtong Zhang \email shangtong@virginia.edu \\
      \addr University of Virginia
      \AND
      \name Lu Feng \email lu.feng@virginia.edu \\
      \addr University of Virginia}
\else
\author{Anonymous authors}
\fi

\begin{document}

\maketitle
\ifarxiv
{\renewcommand{\thefootnote}{}\footnotetext{$^{*}$Equal contribution.}}
\fi

\begin{abstract}
Safe in-context reinforcement learning (Safe ICRL) asks a policy with fixed parameters to improve on an unseen constrained task from its interaction history while aiming to keep its cost within a budget. Prior work studied Safe ICRL on five benchmarks in three families: a grid world with a hidden goal and obstacles (DarkRoom), continuous navigation where goals and hazards stay invisible until reached (PointGoal, CarGoal), and locomotion with a hidden velocity limit (HalfCheetah, Ant). Safe Algorithm Distillation (Safe AD), trained on the learning histories of a safe RL algorithm, has shown adaptation failures in prior evaluations~\citep{moeini2026safeincontextreinforcementlearning, kwon2026latentqbarriershieldingsafe}. For example, in the locomotion tasks its return stays below zero, and it rarely reaches the goal in CarGoal. We revisit two parts of Safe AD. First, we distill a budget-aware teacher with two phases. It searches for the hidden task, exploring more when more budget remains. After its first reward or cost, an expert policy exploits the task within the remaining budget. We collect the teacher's histories under fixed or dynamic budgets. Dynamic budgets increase the occupancy entropy of the return and budget inputs in every environment, and at test time they improve return and lower cost in DarkRoom and HalfCheetah. Second, we propose A-Lag, a test-time adjustment that reweights the distilled policy's actions by their predicted cost, with an adaptive Lagrange multiplier that is updated from the costs observed. In our experiments, A-Lag had lower mean episode cost than Safe AD in all five environments, with lower mean final return in four.
\end{abstract}

\section{Introduction}
In-context reinforcement learning (ICRL) adapts to a new task through the interaction history, without updating model parameters~\citep{moeini2025survey, run2026incontextreinforcementlearningnonstationarity}. When the test task differs from the training tasks, this adaptation poses a safety-generalization problem: changes in dynamics or task conditions may remain partly unobserved, degrade policy generalization, and create safety risks~\citep{kirk2023survey,benjamins2023contextualize, kwon2026safety}. 
Safe ICRL extends ICRL to this setting by asking the policy to improve its task behavior while controlling cost. Transformer policies can learn in-context adaptation for reward-only objectives~\citep{xu2022prompting,laskin2023incontext,dpt2023superivsed,raparthy2023generalization,huang2024context,song2026reward},
and prior work formalized Safe ICRL with constrained Markov decision processes~\citep{moeini2026safeincontextreinforcementlearning}. Algorithm Distillation (AD) trains a Transformer on the learning histories of an RL algorithm~\citep{laskin2023incontext}. Safe AD,
the AD baseline of that work, distills the histories of a safe RL algorithm such as PPO-Lagrangian~\citep{moeini2026safeincontextreinforcementlearning} and has shown adaptation failures on some of these benchmarks. In HalfCheetah and Ant its return stays below zero in every in-context episode, and in CarGoal, it rarely reaches the goal (Appendix~\ref{apx:ppolag}). This motivates the central question of the paper:
\emph{can Safe AD adapt to unseen tasks at lower cost if it learns from a budget-aware teacher and adjusts to the budget explicitly at test time?}

We study two parts of Safe AD, the data it learns from and how it uses the budget at deployment. In safe RL, the learning histories usually come from a collector trained under one fixed cost limit~\citep{achiam2017constrained,ray2019benchmarking}. Safe AD distills the learning histories. These histories show policy improvement, but they are not built to show how behavior should change with the budget or how to search for an unknown task at low cost. We instead distill a teacher whose histories show the behavior we want from the student. It searches for the hidden task while the budget allows it and then switches to an expert policy that exploits what it found with the best behavior that fits the remaining budget, similar to expert-data pretraining for ICRL~\citep{dpt2023superivsed}. At deployment, Safe AD takes the remaining cost budget (cost-to-go, CTG) as an input at every step, but how it reacts to that input is learned in training. For example, if an episode overspends its budget, Safe AD can adjust the next episode only in the way it learned during training, through the history in its context.

To address this, we propose Action-level Lagrangian (A-Lag), which adds an explicit correction while leaving the trained weights unchanged. 
Its Lagrange multiplier penalizes the predicted cost of each action inside the distilled policy rather than a training objective. 
A-Lag keeps Safe AD's Transformer, its return-to-go (RTG) and CTG inputs, and its behavior-cloning objective, and it trains one extra head that predicts the immediate cost of an action from the policy's representation. 
At deployment the distilled policy is reweighted by the predicted cost of each action, scaled by a multiplier $\lambda$, so that actions with high predicted cost become less likely. The multiplier is updated between episodes by a dual-ascent step on the realized episode cost relative to the budget. A-Lag therefore adapts in two ways, through its context as Safe AD does and through the multiplier, which responds explicitly to observed costs relative to the budget. Our experiments use the five Safe ICRL benchmarks of~\citet{moeini2026safeincontextreinforcementlearning}: a grid world with a
hidden goal and obstacles (DarkRoom), continuous navigation where goals and hazards stay invisible until reached (PointGoal, CarGoal), and locomotion with a hidden velocity limit (HalfCheetah, Ant). Our contributions are the following.
\begin{tightitems}
\item \textbf{Budget-aware teacher.} We propose distilling a budget-aware teacher that searches for the hidden task and then exploits it. Collecting its histories under dynamic budgets makes the return and budget inputs more diverse by occupancy entropy (Table~\ref{tab:coverage}).

\item \textbf{A-Lag.} We propose a Lagrange multiplier that reweights actions by predicted cost and is updated from observed episode costs at test time.
In our experiments, A-Lag had lower mean episode cost than Safe AD in all five environments, with lower mean final return in four (Section~\ref{sec:rq2}).
\end{tightitems}

\section{Background}
\label{sec:background}

\paragraph{Constrained Markov Decision Process.}
We model each task as a finite-horizon Constrained Markov Decision Process (CMDP)
$\mathcal{M} = (\fS, \fA, r, c, p, p_0, T, \delta)$, where $c:\fS\times\fA\to\mathbb{R}$ is the cost function and $\delta\ge0$ is the safety budget \citep{Altman2021}.
At time step $t$, the agent observes state $S_t\in\fS$, takes action $A_t\in\fA$ according to policy $\pi(\cdot\,|\, S_t)$, receives reward $R_{t+1}\doteq r(S_t,A_t)$ and cost $C_{t+1}\doteq c(S_t,A_t)$, and transitions to $S_{t+1}\sim p(\cdot\,|\, S_t,A_t)$.
An episode is the trajectory $\tau = (S_0,A_0,R_1,C_1,S_1,\dots,S_{T-1},A_{T-1},R_T,C_T),$
with return $G(\tau)\doteq\sum_{t=1}^T R_t$ and episode cost $G_c(\tau)\doteq\sum_{t=1}^T C_t$.
The objective is to maximize $\mathbb{E}_{\tau\sim\pi}[G(\tau)]$ subject to $\mathbb{E}_{\tau\sim\pi}[G_c(\tau)]\le\delta$.

\paragraph{In-Context Reinforcement Learning.} In-context reinforcement learning (ICRL) pretrains a transformer policy $\pi_{\theta}$ on a static dataset of logged trajectories and then adapts to a new task by conditioning on its growing interaction history, without gradient updates at test time~\citep{laskin2023incontext, kirsch2023towards}.
The parameters $\theta$ are learned once during pretraining and then fixed. At test time the policy conditions on context $H_t^k$ at each step $t$ of episode $k$: $\pi_{\theta}(\cdot \,|\, S_t^k, H_t^k)$. A common choice for the context is the history of observed transitions: $H_t^k \doteq (\tau_1,\dots,\tau_{k-1},\tau_k^t),$
where $\tau_k^t \doteq (S_0^k, A_0^k, R_1^k, C_1^k, S_1^k, \dots, S_{t-1}^k, A_{t-1}^k, R_t^k, C_t^k)$ is the observed prefix of the current episode up to time $t$, with per-step costs included only in the constrained setting. We denote the future realized reward and cost sums by
$
G_{r,t}(\tau) \doteq \sum_{i=t+1}^{T} R_{i}, ~
G_{c,t}(\tau) \doteq \sum_{i=t+1}^{T} C_{i},
$ respectively. 
We use $\mathrm{RTG}_t$ for target rewards, realized in training and user-specified at deployment, and $\mathrm{CTG}_t$ for cost budgets.
This adaptation is often explained as the network running an RL algorithm in its forward pass, which improves as the context grows~\citep{laskin2023incontext,kirsch2023towards,lin2024transformers,wang2025transformers,wang2025towards}.

\paragraph{Safe In-Context Reinforcement Learning.}
Safe ICRL studies the safety of adaptation to unseen tasks~\citep{moeini2026safeincontextreinforcementlearning}. The goal of Safe ICRL is to pretrain parameters $\theta$ such that, when the policy $\pi_{\theta}$ is deployed on unseen CMDPs at test time, performance improves across episodes while satisfying the safety budget $\delta$.
Formally, parameters $\theta$ are learned on training CMDPs.  On a fixed unseen CMDP, the resulting policy is evaluated over episodes $\{\tau_k\}_{k=1}^K$ by
$\E_{\pi_{\theta}}\!\left[\sum_{k=1}^K G(\tau_k)\right]
~
\text{s.t.}
~
\forall k,\;
\E_{\pi_{\theta}}\!\left[G_c(\tau_k)\right] \le \delta.$ 
Since the CMDP and $\theta$ stay fixed during this evaluation, any adaptation comes from the growing context $H_t^k$.

\section{Teacher Data for Safe Algorithm Distillation}
\label{sec:data}
Safe AD can only learn the behavior its histories show. The learning histories of a safe RL algorithm under one fixed cost limit show a policy improving, but not how to search for a hidden task at low cost or how to act when less budget remains. To address this, we collect histories with the two-phase teacher of Section~\ref{sec:teacher}, under the fixed- and dynamic-budget schemes defined below. 

\subsection{Dynamic-Budget Collection}
\label{sec:budgets}
Each training history consists of $K$ in-context episodes on one training task, generated by a teacher policy $\pi_{\mathrm T}$. The budget
$B_0$, the $\delta$ of Section~\ref{sec:background}, is sampled once per history, and the remaining budget $B_t^k$ restarts from it in every episode:
\begin{equation}
\begin{aligned}
B_0&\sim\mathcal D_{\mathrm{budget}}, & B_0^k&=B_0,\\
A_t^k&\sim\pi_{\mathrm T}(\cdot\mid S_t^k,H_t^k,B_t^k), &
B_{t+1}^k&=\max\bigl(0,\,B_t^k-C_{t+1}^k\bigr),
\end{aligned}
\label{eq:collection}
\end{equation}
where $k\in\{1,\ldots,K\}$ indexes in-context episodes, and $S_t^k$ and $H_t^k$ are defined in Section~\ref{sec:background}.
In the \emph{dynamic-budget} collection, $\mathcal D_{\mathrm{budget}}=\mathcal U[B_{\min},B_{\max}]$ over the environment's budget range. 
In the \emph{fixed-budget} collection, every history uses one budget, $B_0=B_{\max}$. Prior Safe ICRL work used fixed cost limits, with the learning histories of a safe RL algorithm such as PPO-Lagrangian~\citep{moeini2026safeincontextreinforcementlearning}. 
Fixed-budget histories still differ in realized cost, since how much the teacher explores depends on the task, but they never show how behavior should change when the starting budget is small. Each logged step carries state, action, reward, cost, $\mathrm{RTG}_t^k=G_{r,t}(\tau_k)$ and $\mathrm{CTG}_t^k=B_t^k$, and the executed action is the training target. RTG is the reward still to come in episode $\tau_k$, while CTG is the remaining budget, not the cost still to come.

\subsection{The Teacher}
\label{sec:teacher}
The teacher does not see the task (the goal location or the velocity threshold)
or the hazard layout. It decides only from the inputs the student also sees: observations, rewards, costs and episode boundaries. Writing $H$ for the history $H_t^k$ together with the
current state $S_t^k$, and $B$ for $B_t^k$, it acts in two phases:
\begin{equation}
\pi_{\mathrm T}(\cdot\mid H,B)=
\begin{cases}
\mathrm{Search}(H,B), & \text{until the first reward (goal tasks) or cost (velocity tasks) in $H$},\\[2pt]
\mathrm{Expert}_q(H,B), & \text{afterwards}.
\end{cases}
\label{eq:teacher}
\end{equation}
Search explores more when more budget remains and avoids places where it has already paid cost. After this switching event, the expert of quality $q$
exploits it with the best behavior it knows. In DarkRoom, the expert follows the
shortest known path with no observed cost, which fits any remaining budget. In the navigation tasks, it drives to the goal position estimated from its history. In the velocity tasks, it runs at the fastest speed whose cost fits $B$. In the navigation and velocity tasks, the expert decides where to go or how fast to run. Then a PPO policy, trained to reach a given position or hold a given speed, produces the actions accordingly. $\mathrm{Search}$, $\mathrm{Expert}_q$ and the event that switches between them depend on the environment (Appendix~\ref{apx:teacher_data} and Table~\ref{tab:teacher}). 

\subsection{Expert Quality}
\label{sec:quality}
$\mathrm{Expert}_1$ is our highest-return expert in each environment. Like every
expert, it uses only behavior whose known cost fits the remaining budget. In DarkRoom it takes the shortest known cost-free path. In the other environments it runs the PPO controller checkpoint with the best validation score (Appendix~\ref{apx:teacher_data}). The quality $q$ of an expert is its return after the switching event, as a fraction of $\mathrm{Expert}_1$'s,
\begin{equation}
q=\frac{\mathbb E\bigl[\bar G_{\mathrm{post}}\bigr]_{\mathrm{Expert}_q}}
{\mathbb E\bigl[\bar G_{\mathrm{post}}\bigr]_{\mathrm{Expert}_1}},
\label{eq:quality}
\end{equation}
where $\bar G_{\mathrm{post}}$ is the mean return of in-context episodes that follow the one with the switching event. We collect histories at $q=1$ and at $q\approx0.5$ and train on both, so the student sees both strong and weak experts. The weaker expert is an earlier checkpoint of the same controller or, in DarkRoom, $\mathrm{Expert}_1$ mixed with random actions at a rate chosen to give $q\approx0.5$. The stronger expert is also the safer one. In the dynamic-budget histories, its violation rate and cost are lower than the weaker expert's in every environment (Appendix~\ref{apx:teacher_data}).

\section{A-Lag: In-Context Dual Adaptation}
\label{sec:architectures}
A-Lag, short for action-level Lagrangian, penalizes the predicted cost of each
action with a Lagrange multiplier and updates that multiplier during deployment.
It combines the Safe AD sequence model, trained as in Safe AD
(Section~\ref{sec:causal_sequence}), a cost head that defines a family of
cost-reweighted policies (Section~\ref{sec:alag_head}), and a deployment-time
update of the multiplier from realized costs
(Section~\ref{sec:dual_adaptation}). Apart from the growing context, only this
update changes how the trained network acts.

\subsection{Sequence Model}
\label{sec:causal_sequence}
We model Safe ICRL as causal action prediction over histories of states,
actions, rewards and
costs~\citep{chen2021decision,laskin2023incontext,xu2022prompting,moeini2026safeincontextreinforcementlearning}.
Each completed transition adds four positions: state, action, reward and cost.
States, actions, rewards and costs have different input domains, so separate
encoders $E_{\mathtt{s}}$, $E_{\mathtt{a}}$, $E_{\mathtt{r}}$ and
$E_{\mathtt{c}}$ map them into a shared embedding space, and further encoders
$E_{\mathrm{RTG}}$ and $E_{\mathrm{CTG}}$ map the two control inputs into the
same space. Scalars are encoded with the signed logarithm
\begin{equation}
\psi(z)=\operatorname{sign}(z)\log(1+|z|),
\label{eq:signed_log}
\end{equation}
which preserves sign and order and reduces scale differences across
environments. Let $p=(k,t)$ identify step $t$ of episode $k$, with
$R_{p+1}\doteq R_{t+1}^k$ and $C_{p+1}\doteq C_{t+1}^k$. As in Safe AD, the
control inputs enter the state position:
\begin{equation}
\begin{aligned}
\mathbf e_{p,\mathtt{s}}
&=E_{\mathtt{s}}(S_p)
+E_{\mathrm{RTG}}\!\left(\psi(\mathrm{RTG}_p)\right)
+E_{\mathrm{CTG}}\!\left(\psi(\mathrm{CTG}_p)\right),\\
\mathbf e_{p,\mathtt{a}}&=E_{\mathtt{a}}(A_p),\quad
\mathbf e_{p,\mathtt{r}}=E_{\mathtt{r}}\!\left(\psi(R_{p+1})\right),\quad
\mathbf e_{p,\mathtt{c}}=E_{\mathtt{c}}\!\left(\psi(C_{p+1})\right).
\end{aligned}
\label{eq:field_content_embedding}
\end{equation}
A sampled training window $\boldsymbol\tau$ is a consecutive window of these positions from one
$K$-episode history. The state position precedes the action, reward and cost
positions of its transition, so under causal masking the prediction of $A_p$
cannot see that action or its outcome. The action head reads the last-layer
representation $\mathbf u_p$ at the state position of step $p$ in an
$L$-layer Transformer and defines the base policy $\pi_\theta(\cdot\mid\mathbf
u_p)$: a categorical distribution with logits $\ell(\mathbf u_p,\cdot)$ for
discrete actions, or a diagonal Gaussian with mean $\mu_d(\mathbf u_p)$ and
variance $v_d(\mathbf u_p)$ in action dimension $d$ for continuous actions.
Training minimizes the negative log-likelihood of the logged action,
\begin{equation}
\mathcal L_{\mathrm{act}}
=\mathbb E_{\boldsymbol\tau}\!\left[-\sum_p m_p\log
\pi_\theta\bigl(A_p\mid\mathbf u_p\bigr)\right],
\end{equation}
where $m_p$ masks padded positions. This is Safe AD's model and objective.

\subsection{Cost Head and the Cost-Reweighted Policy}
\label{sec:alag_head}
A second head $\hat c(\mathbf u_p,\cdot)$ scores each action by its predicted
immediate cost, scaled by $c_0$ (Eq.~\eqref{eq:alag_cost_loss}). For discrete actions it outputs one score per action. For
continuous actions it is a quadratic in each action dimension, $\hat c(\mathbf u,A)=b(\mathbf u)+\sum_d\bigl(g_d(\mathbf u)A_d
+\tfrac12\kappa_d(\mathbf u)A_d^2\bigr)$ with $\kappa_d\ge0$. For a multiplier
$\lambda$ the head defines the cost-reweighted policy
\begin{equation}
\pi_\lambda(A\mid\mathbf u_p)\;\propto\;
\pi_\theta(A\mid\mathbf u_p)\,
\exp\bigl(-\lambda\,\hat c(\mathbf u_p,A)\bigr).
\label{eq:alag_policy}
\end{equation}
For discrete actions the reweighted logits are
$\ell(\mathbf u_p,A)-\lambda\,\hat c(\mathbf u_p,A)$. For continuous actions the
cost-reweighted policy is Gaussian in closed form, with variance
$(v_d^{-1}+\lambda\kappa_d)^{-1}$ and mean
$(v_d^{-1}\mu_d-\lambda g_d)/(v_d^{-1}+\lambda\kappa_d)$ in dimension $d$. A
positive $\lambda$ shifts probability toward actions with low predicted cost,
and $\pi_0=\pi_\theta$.

\paragraph{Training.} The cost head is fitted by regression to the
scaled immediate cost of the executed action, maintaining the policy representation fixed for this loss:
\begin{equation}
\mathcal L_{\mathrm{cost}}
=\mathbb E_{\boldsymbol\tau}\!\left[\sum_p m_p
\bigl(\hat c(\operatorname{sg}(\mathbf u_p),A_p)-C_{p+1}/c_0\bigr)^2\right],
\label{eq:alag_cost_loss}
\end{equation}
where $\operatorname{sg}$ blocks the cost-loss gradient into the backbone and $c_0>0$ is the standard deviation of the immediate cost in the
environment's training data. We minimize
$\mathcal L_{\mathrm{act}}+\eta_C\,\mathcal L_{\mathrm{cost}}$ with $\lambda=0$
throughout training. The cost loss trains the cost head, while the action loss trains the backbone and action head.

\subsection{In-Context Dual Adaptation}
\label{sec:dual_adaptation}
In a deployment of $K$ episodes with budget $B_0$, A-Lag keeps one multiplier
$\lambda_k$ for each episode $k$. It starts at $\lambda_1=0$, and after episode
$k$ it is updated from the realized cost of that episode,
\begin{equation}
\lambda_{k+1}=\Pi_{[0,\lambda_{\max}]}\!\left(
\lambda_k+\eta\,\frac{G_c(\tau_k)-B_0}{B_{\max}}\right),
\label{eq:dual_update}
\end{equation}
where $\eta>0$ is a step size, $B_{\max}$ the upper end of the environment's budget range (Section~\ref{sec:budgets}), used as a fixed scale, and $\Pi$ the projection onto $[0,\lambda_{\max}]$. Within episode $k$ every action is sampled from $\pi_{\lambda_k}(\cdot\mid\mathbf u_p)$. The context and
the RTG and CTG inputs are those of Safe AD, and $\lambda$ restarts at zero in the next deployment. Episode costs above the budget raise $\lambda$ and
make the next episode more conservative. Costs below it lower $\lambda$ again, at most down to $0$, where A-Lag acts as its base policy.

Eq.~\eqref{eq:dual_update} has the form of projected dual subgradient ascent on the Lagrangian of the episode constraint
$\mathbb E[G_c(\tau_k)]\le B_0$, with the realized episode cost as a one-sample estimate of the constraint value. Lagrangian safe RL applies this
update to its multiplier during training~\citep{Altman2021,ray2019benchmarking}. A-Lag uses it during a single test deployment of a distilled policy, to tune the multiplier from the costs the agent incurs, with no retraining and no choice of $\lambda$ per task. Only the upper bound $\lambda_{\max}$ is set per environment, in advance on training tasks by a fixed rule. The step size, in units of $\lambda_{\max}$, is the same in every environment (Appendix~\ref{apx:controller}).

\subsection{Properties}
\label{sec:alag_theory}
For a fixed representation $\mathbf u$, let
$Z_\lambda=\mathbb E_{A\sim\pi_\theta}[\exp(-\lambda \hat c(\mathbf u,A))]$ be the
normalizer of $\pi_\lambda$. It is finite for every $\lambda\ge0$, both for
finite action sets and in the Gaussian case. For every $\lambda\ge0$, $\pi_\lambda$ is the unique minimizer, over
action distributions $\pi$ with finite objective, of $\mathrm{KL}\bigl(\pi\,\|\,\pi_\theta(\cdot\mid\mathbf u)\bigr)
+\lambda\,\mathbb E_{A\sim\pi}\bigl[\hat c(\mathbf u,A)\bigr],$ because this objective is equal to $\mathrm{KL}(\pi\,\|\,\pi_\lambda)-\log
Z_\lambda$. 
For $\lambda\ge0$, the cost-reweighted policy is the solution of a KL-regularized Lagrangian relaxation in which
$\lambda$ penalizes the predicted cost and the distilled policy plays the role of the unconstrained policy. The next proposition shows that raising the multiplier never raises the expected predicted cost.

\begin{proposition}[Monotone predicted cost]
\label{prop:alag_mono}
Assume the history in the context is fixed, so that $\mathbf u_p$ does not change.
Then for every $\lambda\ge0$,
\begin{equation}
\frac{d}{d\lambda}\,\mathbb E_{A\sim\pi_\lambda}\bigl[\hat c(\mathbf u_p,A)\bigr]
=-\operatorname{Var}_{A\sim\pi_\lambda}\bigl(\hat c(\mathbf u_p,A)\bigr)\le 0 .
\label{eq:alag_dvar}
\end{equation}
\end{proposition}
Appendix~\ref{apx:proofs} gives the proof and the derivation of the KL
characterization above.
By Eq.~\eqref{eq:dual_update}, an episode whose cost is above the budget $B_0$ does not decrease $\lambda$, and by
Proposition~\ref{prop:alag_mono}, at every fixed history the next episode's expected predicted cost per step then does not increase. 
An episode below the budget does not increase $\lambda$, so the expected predicted cost does not decrease. 
This holds only for the predicted cost with the context fixed. It does not mean that the actual episode cost falls as $\lambda$ grows, since different actions lead to a different context, and it does not guarantee that the budget is met. The cost head predicts only the cost of the next step, so it cannot predict later costs.

\section{Experiments}
\label{sec:experiments}
We ask three questions. \textbf{RQ1}: do dynamic budgets, which make the teacher data cover more
diverse return and budget inputs, improve test return and cost over a fixed
budget? \textbf{RQ2}: does
A-Lag improve on the return and cost of Safe AD on unseen test
tasks, and how does CDT compare? \textbf{RQ3}: how often and by how much does
each method go over its budget?

\subsection{Experimental Setup}
We use five environments derived from the Safe ICRL benchmarks of~\citet{moeini2026safeincontextreinforcementlearning}.
Each has separate training and test tasks. Appendix~\ref{apx:envs} gives the layouts and the sizes of the splits.
\begin{itemize}
\item \textbf{DarkRoom} is a partially observed grid world with a hidden goal
and hidden obstacles. Training goals and obstacles are center-oriented, while test goals and obstacles are edge-oriented.
\item \textbf{PointGoal} and \textbf{CarGoal} are continuous navigation tasks with a broken lidar, so the agent cannot sense goals or hazards until it reaches or touches them. Their training layouts are center-oriented and their test layouts are edge-oriented.
\item \textbf{HalfCheetahVelocity} and \textbf{AntVelocity}, below called HalfCheetah and Ant or the velocity tasks, are locomotion tasks in which cost is incurred above a hidden velocity limit that differs between tasks~\citep{ji2023safety}. The test velocity limits are unseen.
\end{itemize}

\paragraph{Models.} Safe AD~\citep{moeini2026safeincontextreinforcementlearning}
and A-Lag are both trained from scratch using the backbone of Section~\ref{sec:causal_sequence}. Unless otherwise stated, Safe AD here is trained on the teacher data, not PPO-Lagrangian histories. The main models use 32 histories per task. We report the 50\% mixture. A-Lag adds the cost head and the deployment-time update. CDT~\citep{liu2023cdt} is
an offline safe-RL sequence policy that conditions on RTG and CTG. All models
train for 100{,}000 updates with three random seeds and the same
per-environment architecture. We also show two safe meta-RL baselines from a preliminary study,
MAML with a cost penalty~\citep{finn2017model} and SafeMeta~\citep{xu2025efficient}, which adapt by updating policy parameters on each new task. We do not compare with SCARED~\citep{moeini2026safeincontextreinforcementlearning} or latent Q-barrier shielding~\citep{kwon2026latentqbarriershieldingsafe} because they are pretrained online, interacting with the training environments under their own objectives. Safe AD and A-Lag instead learn offline from fixed teacher histories, and our question is what the data and a deployment-time multiplier do for AD itself.

\paragraph{Evaluation.} A deployment runs one test task for $K$ episodes with fixed parameters while the context grows. History is retained across episode resets within a deployment and cleared between deployments. $K$ is 50 in DarkRoom, 20 in PointGoal and CarGoal and 10 in the velocity tasks. Each deployment samples its initial RTG uniformly between two values fixed from the training data, the lowest and the median over training tasks of the teacher's best in-context episode return, and its budget $B_0$ uniformly over the training budget range. Both stay fixed within the deployment. Every test task receives 15 deployments, and the draws are shared by all methods. We report four quantities, averaged over draws, tasks and seeds. The final return is the mean return of the last three episodes. In PointGoal and CarGoal the reward is sparse. The episode return is 1 if the goal
is reached and 0 otherwise, so there the return is the success rate. The cost is the mean episode cost $G_c$ over all $K$ episodes. The violation rate is the fraction of these episodes whose cost exceeds $B_0$, and the excess cost is their mean of $\max(0,G_c-B_0)$.

Each curve point is the mean of that episode over all deployments, test tasks and seeds, so it pools deployments with different budgets. Table entries average these over the episodes. The
shaded bands in the figures show $\pm2$ standard errors over deployments. Text comparisons pair methods on the same seed and deployment draw. Main-text comparisons describe observed means. Appendix~\ref{apx:seeds} gives per-seed results and intervals over test tasks.

\paragraph{Tuning A-Lag.} We set $\lambda_{\max}$ per environment from a sweep
of fixed multipliers on training tasks (Figure~\ref{fig:grid}). It is the
largest tested value up to which the mean final return stays above 75\% of that
at $\lambda=0$. The step size $\eta=\lambda_{\max}$ was
chosen once in DarkRoom and is used in every environment
(Appendix~\ref{apx:controller}). The primary
comparison is A-Lag against Safe AD on the test tasks. In no environment does
A-Lag have both a lower mean violation rate and a higher mean final return (Table~\ref{tab:main}).

\subsection{RQ1: Fixed vs.\ Dynamic Budgets in the Teacher Data}
\label{sec:rq1}
We train Safe AD and A-Lag on the fixed-budget histories, in which
$B_0=B_{\max}$, and compare them with the main models, which are trained on the
dynamic-budget histories. Both use the same training settings, so the two
differ only in the data.

\paragraph{Why coverage matters.} Safe AD and A-Lag choose each action from the history and two control inputs, the return still to be reached (RTG) and the budget still left (CTG). Each deployment samples a new budget, and the remaining budget falls as cost is incurred, so at test time the model sees many (RTG, CTG) pairs. It is most likely to act well at pairs that occur in its training data. Under a fixed budget every episode starts with the remaining budget at $B_{\max}$, and a small remaining budget appears only after cost has been incurred. 
Table~\ref{tab:coverage} measures how diverse the two kinds of histories are, and Figure~\ref{fig:coverage} in
Appendix~\ref{apx:coverage} shows where their steps fall. More occupied cells mean that the histories reach more (RTG, CTG) pairs, and a higher entropy means that their steps spread more evenly over these pairs instead of concentrating on a few. On a $20\times20$ grid over RTG and CTG, the dynamic-budget steps occupy all 400 cells in DarkRoom and the velocity tasks, while the fixed-budget steps occupy 179 in DarkRoom. The dynamic-budget histories also have the higher entropy in all five environments.

\begin{figure*}[t]
  \centering
  \includegraphics[width=\textwidth]{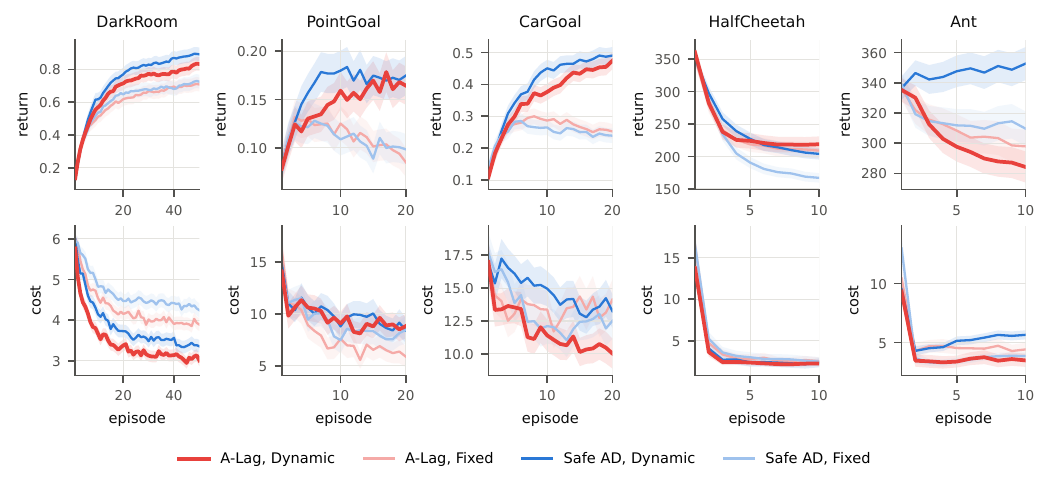}
  \caption{Models trained on dynamic-budget histories (Dynamic) and on fixed-budget histories (Fixed), on the test tasks. 
  Each curve shows the mean episode return or cost over three seeds, with a shaded band of $\pm2$ standard errors.
  With dynamic budgets, A-Lag has a lower cost in four of the five environments,
  all but PointGoal, and Safe AD has a higher return in all five, compared with fixed budgets.
  }
  \label{fig:rq1_curves}
\end{figure*}

\paragraph{Test results.} Figure~\ref{fig:rq1_curves} shows how the main models
and the fixed-budget models adapt over the episodes of a deployment. In
DarkRoom, fixed-budget data lower the final return by 0.167 for Safe AD and by
0.123 for A-Lag and raise the mean episode cost of both by about 0.8. Both gaps
open early in the deployment and last to its end. In the other environments the
main models return more in every case except A-Lag in Ant. The gap is largest
in PointGoal and CarGoal, where the return of the main models is nearly twice as high.
Outside DarkRoom the costs move in both
directions. They are lower for the main models in HalfCheetah and higher
for both methods in PointGoal. In CarGoal and Ant they are lower for A-Lag and
higher for Safe AD. The entropy is higher in every environment, but return and cost move in the
favorable direction for both methods only in DarkRoom and HalfCheetah.

\begin{table}[t]
\centering
\caption{Coverage of the (RTG, CTG) plane in the teacher data under fixed and dynamic budgets (both quality levels). Each input is rescaled to $[0,1]$ between its near-minimum and near-maximum (the 1st and 99th percentiles over both datasets, with values outside clipped), and the plane is divided into a $20\times20$ grid. Cells: occupied cells out of 400. $\mathrm{Entropy}=-\sum_b p_b\log p_b/\log 400$, where $p_b$ is the share of steps in cell $b$. $\mathrm{Entropy}$ is 1 when the steps spread evenly over all cells (Appendix~\ref{apx:coverage}). Higher values mean more diverse training data. Bold marks the higher value of each pair. In PointGoal and CarGoal the RTG takes only the values 0 and 1, so at most 40 cells can be occupied and $\mathrm{Entropy}$ is at most $\log 40/\log 400\approx0.62$.}
\label{tab:coverage}
\small
\setlength{\tabcolsep}{3pt}
\begin{adjustbox}{max width=\textwidth}
\begin{tabular}{@{}lcccc@{}}
\toprule
& \multicolumn{2}{c}{Cells} & \multicolumn{2}{c}{Entropy} \\
\cmidrule(lr){2-3}\cmidrule(lr){4-5}
Environment & Fixed & Dynamic & Fixed & Dynamic \\
\midrule
DarkRoom & 179 & \textbf{400} & 0.549 & \textbf{0.935} \\
PointGoal & 40 & 40 & 0.346 & \textbf{0.551} \\
CarGoal & 40 & 40 & 0.371 & \textbf{0.560} \\
HalfCheetah & 315 & \textbf{400} & 0.707 & \textbf{0.894} \\
Ant & 279 & \textbf{400} & 0.696 & \textbf{0.906} \\
\bottomrule
\end{tabular}
\end{adjustbox}
\end{table}

\subsection{RQ2: A-Lag vs.\ Safe AD and CDT}
\label{sec:rq2}
\begin{figure*}[t]
  \centering
  \includegraphics[width=\textwidth]{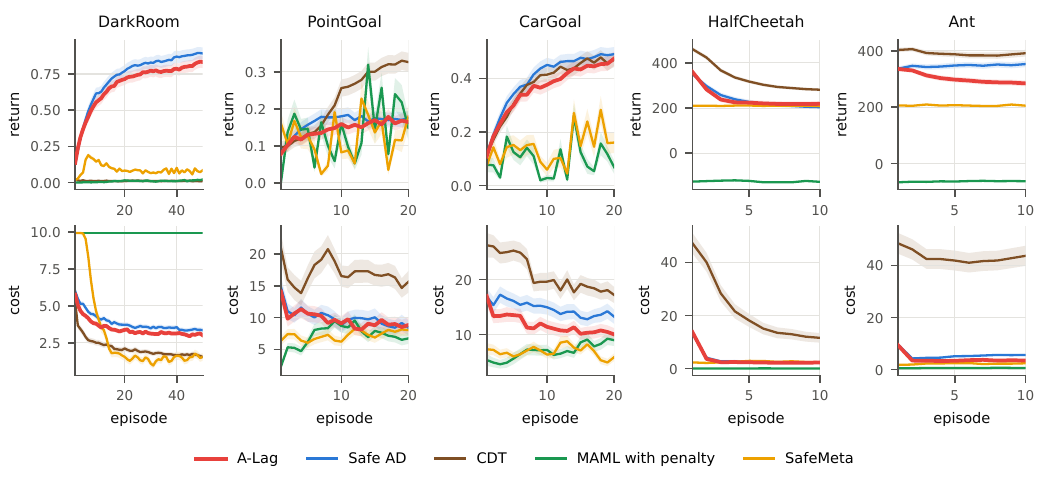}
  \caption{In-context adaptation on the test tasks. Each curve shows the mean
  episode return or cost over three seeds, with a shaded band of $\pm2$ standard errors. A-Lag starts every deployment at $\lambda=0$, and $\lambda$ is adjusted during test time by Eq.~\eqref{eq:dual_update}. In DarkRoom, PointGoal and CarGoal the return of A-Lag and Safe AD rises over the episodes while their cost falls. In HalfCheetah and Ant their cost drops
  sharply after the first episode, and in HalfCheetah the return falls with it. A-Lag's Ant return also declines across episodes. A-Lag has a lower cost than Safe AD in DarkRoom, CarGoal and Ant, at a lower
  return. The mean costs are also slightly lower in PointGoal, at a slightly lower return, and in HalfCheetah, where it ends with a higher return. CDT has a higher cost than both in the navigation and velocity tasks and does not find the goal in DarkRoom.
  }
  \label{fig:test_curves}
\end{figure*}

A-Lag has a lower mean episode cost than Safe AD in all five environments (Table~\ref{tab:main}), with smaller differences in PointGoal and HalfCheetah. The trade-off is a lower mean final return in DarkRoom, PointGoal, CarGoal and Ant. The largest drop is in Ant, 286.4 against 350.9, about 18\%, against 5 to 7\% in the other three. In HalfCheetah A-Lag returns 11.8 more, and averaged over all in-context episodes rather than the last three, the returns of A-Lag and Safe AD are nearly equal, 242.6 and 242.5. Figure~\ref{fig:test_curves} shows the return and the cost over the episodes of a deployment.

Compared with $\lambda=0$, the multiplier lowers mean cost in four environments. In Ant it raises mean cost and excess cost while lowering the violation rate, and A-Lag's lower cost than Safe AD is already present with the multiplier held at zero (Appendix~\ref{apx:tables}).

Table~\ref{tab:mix} (column Main) gives CDT's numbers. In DarkRoom, CDT does not
learn to find the goal, with a final return of 0.010. In the navigation tasks, CDT returns more than Safe AD in
PointGoal, 0.326 against 0.173, and slightly less in CarGoal, 0.468 against
0.489, at a higher mean cost in both, 16.98 and 20.97 against 10.04 and 14.77.
PointGoal return stays below 0.35 for every method, so the small differences
between A-Lag and Safe AD there give little evidence either way. In the velocity tasks, CDT returns more than
Safe AD, 284.9 and 387.0 against 206.7 and 350.9, and its mean cost is 18.4 and
37.7 above Safe AD's.

The two safe meta-RL baselines show little in-context adaptation. In the
velocity tasks their curves are flat. SafeMeta's return stays near 210 at a
mean cost of about 2.5, and MAML with a cost penalty has negative return at a
cost below 1. In DarkRoom and the navigation tasks their return does not rise
steadily over the episodes. These curves come from a different protocol and are shown only for context.

\subsection{RQ3: Budget Violations}
\label{sec:rq3}
\begin{figure*}[t]
  \centering
  \includegraphics[width=\textwidth]{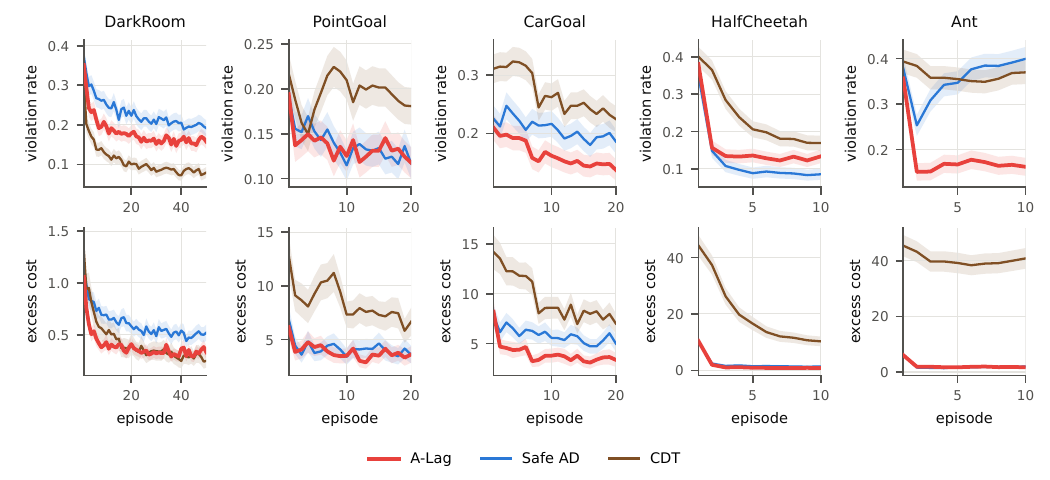}
  \caption{Violation rate and excess cost $\max(0,G_c-B_0)$ over the episodes of
  a deployment on the test tasks. Each curve is the mean over three seeds, with
  a shaded band of $\pm2$ standard errors. Averaged over the episodes, A-Lag has
  the lowest excess cost in DarkRoom, PointGoal, CarGoal and HalfCheetah and is
  close to Safe AD in Ant. In most later DarkRoom episodes CDT, which does not
  find the goal, is lower.
  }
  \label{fig:test_curves_viol}
\end{figure*}

Prior Safe ICRL work evaluates safety by the mean episode cost~\citep{moeini2026safeincontextreinforcementlearning, kwon2026latentqbarriershieldingsafe}. This is not enough, because budgets differ between deployments, so a low mean cost can still hide frequent or large budget violations. In HalfCheetah, for example, A-Lag has a lower mean cost than Safe AD but goes over its budget more often (Table~\ref{tab:main}). Hence, we also report how often a method exceeds its budget and by how much. Figure~\ref{fig:test_curves_viol} shows how often and by how much each method goes over its budget during a deployment. A-Lag's violation rate and excess cost drop after the first episode in every environment. In Ant, Safe AD's violation rate drops at the second episode but rises again over the remaining episodes, while A-Lag's stays low.

A-Lag has a lower mean violation rate than Safe AD in DarkRoom, PointGoal, CarGoal and Ant. The reductions in DarkRoom, CarGoal and Ant are 0.049, 0.043 and 0.172, but as RQ2 showed it also returns less there. In PointGoal the mean violation rates are 0.137 for A-Lag and 0.139 for Safe AD. In HalfCheetah A-Lag violates more often.

Compared with $\lambda=0$ (Appendix~\ref{apx:tables}), the multiplier lowers the mean violation rate in all five environments. The smallest mean reduction is in PointGoal. In HalfCheetah A-Lag at
$\lambda=0$ already violates more often than Safe AD, and the multiplier reduces it.

\paragraph{Why the violation rate and the excess cost disagree.} The excess cost
is the violation rate times the average amount by which a violating episode
exceeds its budget,
\begin{equation}
\mathbb{E}\bigl[\max(0,G_c-B_0)\bigr]=\Pr(G_c>B_0)\,
\mathbb{E}\bigl[G_c-B_0\mid G_c>B_0\bigr].
\end{equation}
A method can therefore go over its budget less often than another and still have the same excess cost, if it exceeds the budget by more each time, and the reverse. In Ant, A-Lag's violation rate is 0.184, about half of Safe AD's 0.356, yet both have an excess cost of about 2.2, because a violating A-Lag
episode exceeds its budget by about 12 on average against 6 for Safe AD. The multiplier adds to this, since compared with $\lambda=0$ it lowers the violation rate from 0.210 to 0.184 but raises the excess cost from 1.56 to 2.23. HalfCheetah goes the other way. There A-Lag goes over the budget more often than Safe AD, 0.158 against 0.123, but by about 13 each time against 20, so its excess cost is 0.47 lower. In the other
three environments the two measures move in the same direction.

In DarkRoom, where it does not find the goal, CDT violates the budget least
often of the three methods, 0.108, and its excess cost of 0.43 lies between
A-Lag's 0.39 and Safe AD's 0.60. In the navigation tasks, CDT goes over the
budget more often than Safe AD, 0.195 and 0.270 against 0.139 and 0.207, with a
higher excess cost, 8.5 and 9.5 against 4.2 and 5.8. In the velocity tasks, it
also violates more often, 0.239 and 0.364 against 0.123 and 0.356, and its
excess cost is 20.2 and 40.5 against Safe AD's 2.5 and 2.2.

\paragraph{Additional analyses.} For Table~\ref{tab:fixed}, we sweep fixed values of $\lambda$ from 0.3 to 1000 on training tasks (Appendix~\ref{apx:controller}). A larger multiplier lowers the return and usually the violation rate, so we compare the adaptive multiplier with the fixed value that has the lowest violation rate among those whose return is not clearly below that at $\lambda=0$. Compared with $\lambda=0$, the multiplier lowers mean cost in every environment but Ant (Appendix~\ref{apx:tables}). Per seed, A-Lag's cost is lower than Safe AD's in all three seeds in CarGoal and Ant and in two of three elsewhere (Appendix~\ref{apx:seeds}). In the expert-share study, A-Lag has a lower mean cost than Safe AD in 16 of 25 combinations of environment and share of $q=1$ histories, so the main result does not hold for every data mixture (Appendix~\ref{apx:mix}). In DarkRoom, dynamic budgets also lower the violation rate, by 0.085 for Safe AD and 0.097 for A-Lag (Appendix~\ref{apx:tables}).

\section{Related Work}
\label{sec:related_work}
\paragraph{In-Context Reinforcement Learning.}
Adapting to new tasks without parameter updates has roots in recurrent
meta-RL~\citep{duan2016rl,wang2016learning,beck2023survey,moeini2025survey}.
Modern ICRL instead pretrains a fixed sequence policy to improve through its
interaction history. Algorithm Distillation learns from complete RL learning
histories~\citep{laskin2023incontext,zisman2024emergence}, whereas supervised
pretraining methods learn from pretrained policies or action
labels~\citep{dpt2023superivsed,xu2022prompting,huang2024context}, and the
Decision Importance Transformer weights the imitation loss by learned advantages
so that it can learn from suboptimal histories~\citep{dong2026incontextreinforcementlearningsuboptimal}. Later work
covers multi-task agents, varied action spaces and in-context Q-learning
\citep{grigsby2024amagoiclr,grigsby2024amagonips,sinii2024variable,polubarov2025vintix,liu2026scalable},
and theory studies when Transformers learn in context
\citep{lin2024transformers,wang2025transformers,wang2025towards}.
Recent analyses connect temporal-difference policy evaluation to specific
Transformer constructions, including softmax attention~\citep{xie2026beyond} and
chain-of-thought generation with linear Transformers~\citep{xie2026convergence}.
These works mainly study reward maximization.

\paragraph{Safe Reinforcement Learning.}
Safe RL commonly uses CMDPs to maximize reward subject to expected-cost
constraints~\citep{garcia2015comprehensive,gu2022review,wachi2024survey}.
Policy optimization~\citep{achiam2017constrained,zhang2020focops,yang2020PCPO,xu2021crpo,yang2022cup,bai2023achieving},
safe exploration~\citep{wachi2020safe,as2025actsafe} and runtime
shielding~\citep{kwon2026adaptiveshieldingsafereinforcement} update a policy or
model online, or filter actions, rather than distilling an adaptation algorithm
into context. Offline sequence policies,
including CDT and safety-conditioned variants
\citep{liu2023cdt,wang2024safe,guo2024temporal}, condition on return and cost but
do not target repeated contextual adaptation to unseen CMDPs. Other work handles
changing budgets within a known task, by conditioning one policy on the cost
threshold~\citep{yao2023constraint} or by inferring trajectories for a budget
set at deployment from offline data~\citep{lin2023safe}.

\paragraph{Safe In-Context and Meta Reinforcement Learning.}
Safe meta-RL adapts to constrained tasks through gradient updates or additional
task-specific data~\citep{luo2021mesa,khattar2023a,guan2024cost,xu2025efficient}.
Safe ICRL instead keeps parameters fixed. SCARED formalizes this setting and
uses budget-sensitive Lagrangian pretraining~\citep{moeini2026safeincontextreinforcementlearning}. More recently,
latent Q-barrier shielding adds an explicit deployment-time filter that
reweights actions from a frozen Safe ICRL policy using predicted future cost
\citep{kwon2026latentqbarriershieldingsafe}. A-Lag also acts at deployment,
but it penalizes only the predicted immediate cost, with one multiplier per
deployment updated from realized episode costs.

\section{Conclusion}
\label{sec:conclusion}
We revisited the histories AD distills and the way the distilled policy uses the
budget, with a budget-aware teacher and A-Lag, a
deployment-time Lagrange multiplier.
A-Lag had a lower mean cost than Safe AD in all five environments, with lower mean final return in four. Against fixed multipliers, the adaptive multiplier had lower mean cost in four environments, but higher cost and return in Ant. Dynamic budgets during data collection gave
higher occupancy entropy of the return and budget inputs in every environment and, in DarkRoom, improved both return and violation rate. Unlike prior Safe ICRL evaluations, which report the mean episode cost, we also report how often and by how much each method exceeds its budget. In HalfCheetah, A-Lag has a lower mean cost than Safe AD but exceeds its budget more often, and in Ant and HalfCheetah the violation rate and the excess cost disagree, so both should be reported. Our study has limits. The cost head predicts only immediate cost and is learned only from the actions the teacher took. The teacher also requires environment-specific search, experts and switching events.

\bibliography{main}
\bibliographystyle{tmlr}

\clearpage

\appendix
\raggedbottom
\section{Environments and Task Splits}\label{apx:envs}
\begin{figure*}[t]
\centering
\begin{tikzpicture}
  \matrix (m) [matrix of nodes,
             nodes={inner xsep=5pt, inner ysep=4pt, outer sep=0, anchor=center},
             row sep=8mm] {
    \includegraphics[width=.24\linewidth]{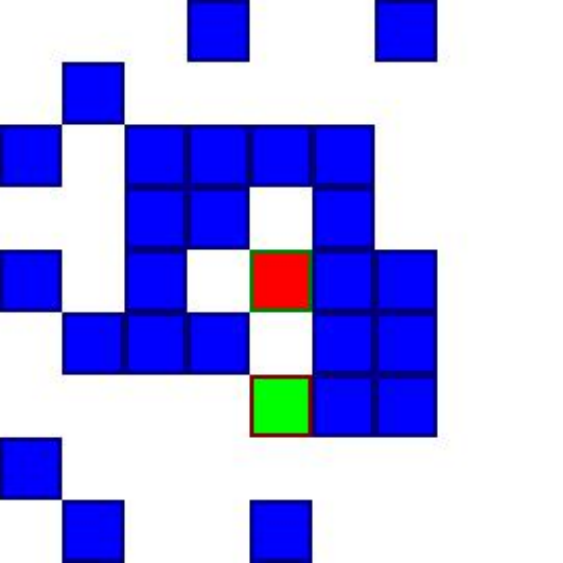} &
    \includegraphics[width=.24\linewidth]{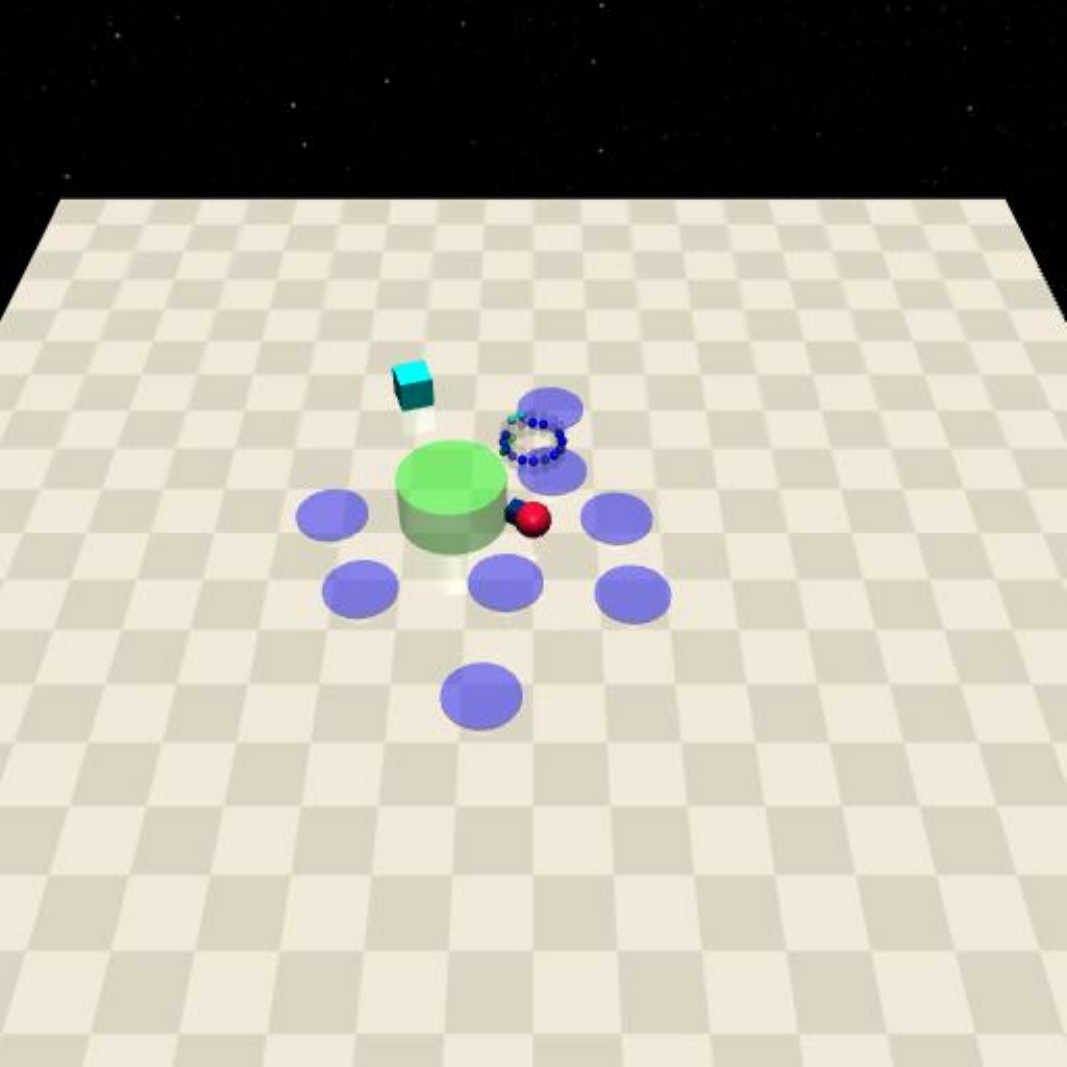} &
    \includegraphics[width=.24\linewidth]{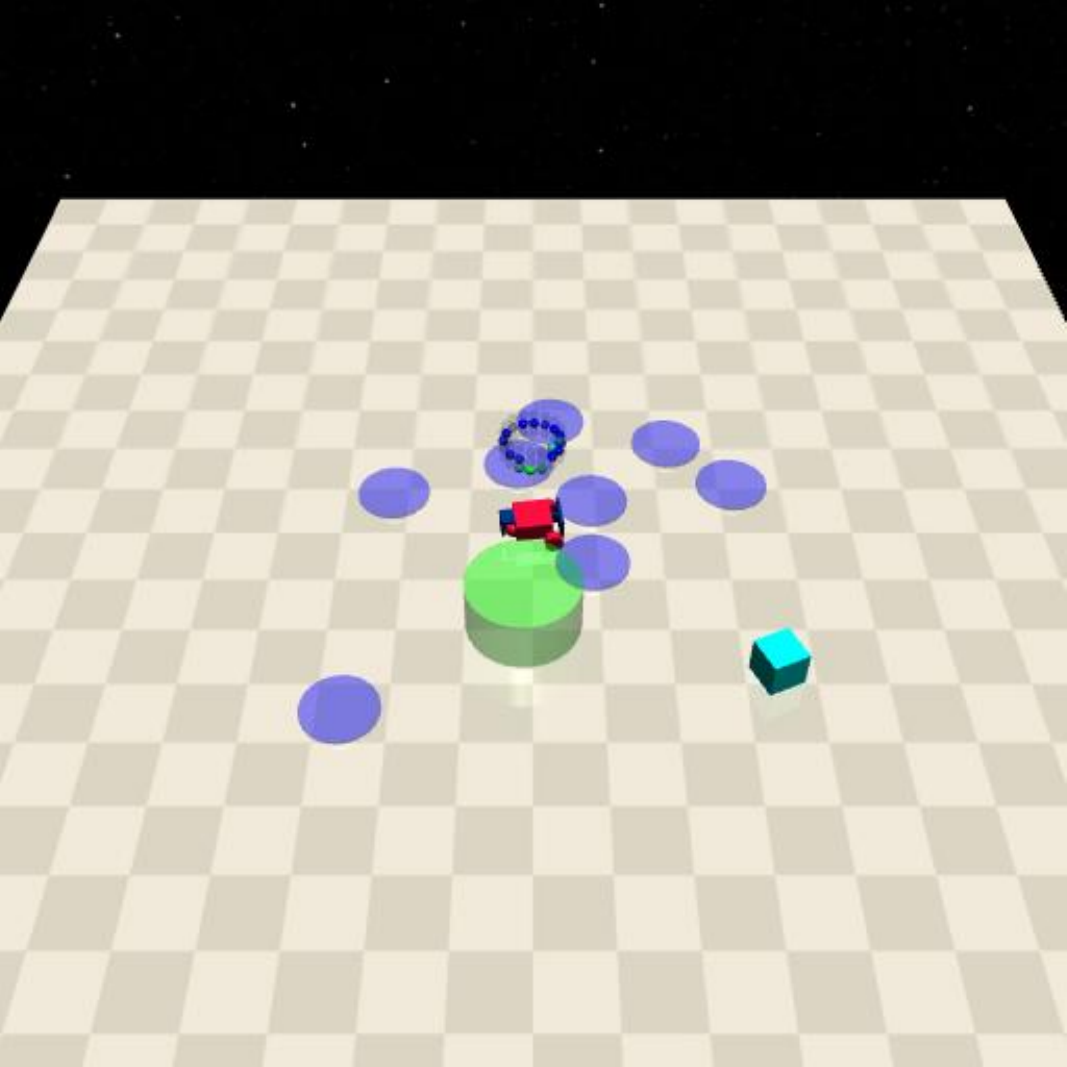} \\
    \includegraphics[width=.24\linewidth]{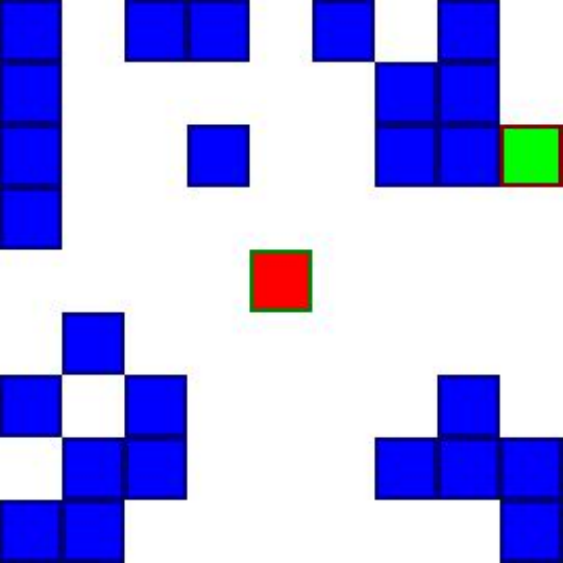} &
    \includegraphics[width=.24\linewidth]{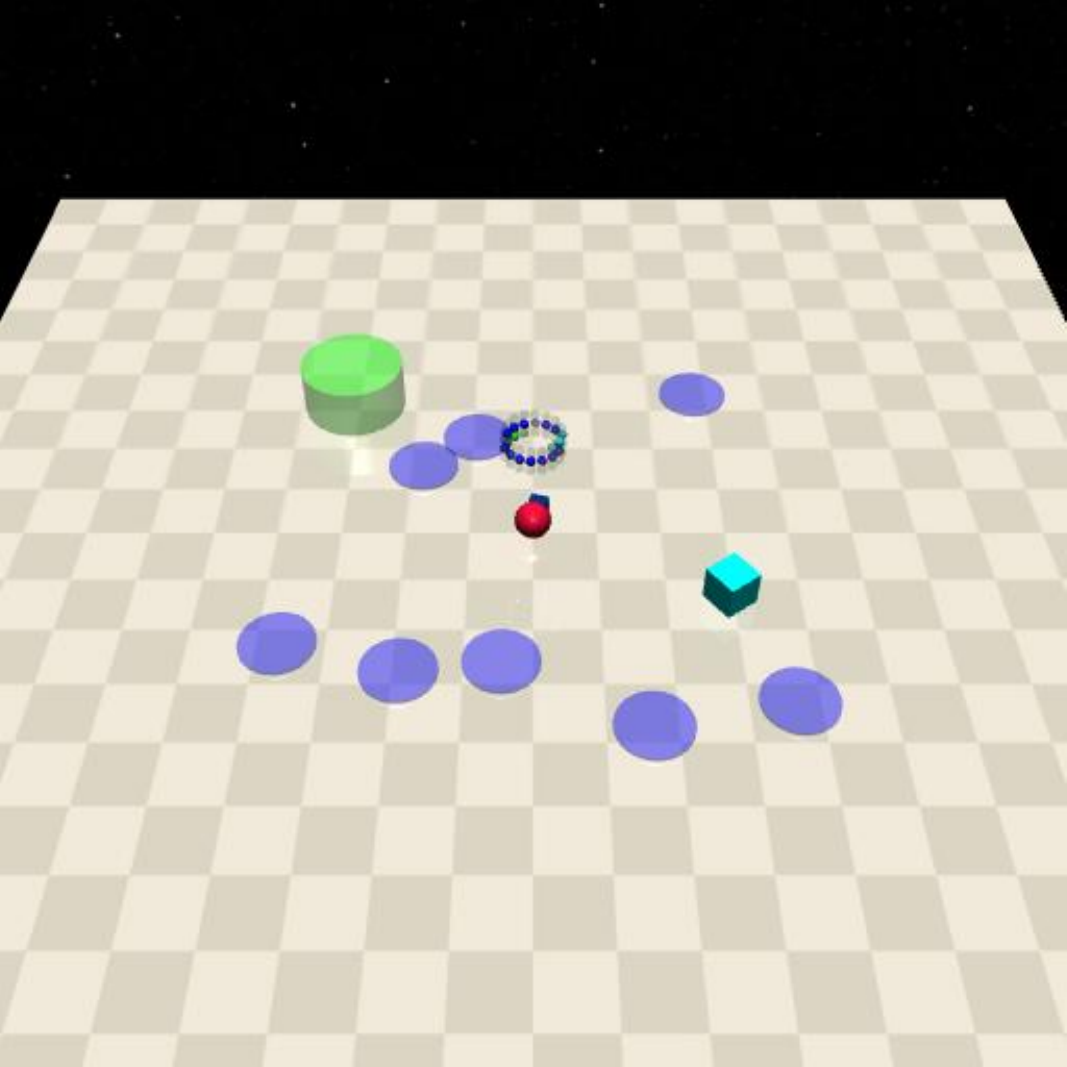} &
    \includegraphics[width=.24\linewidth]{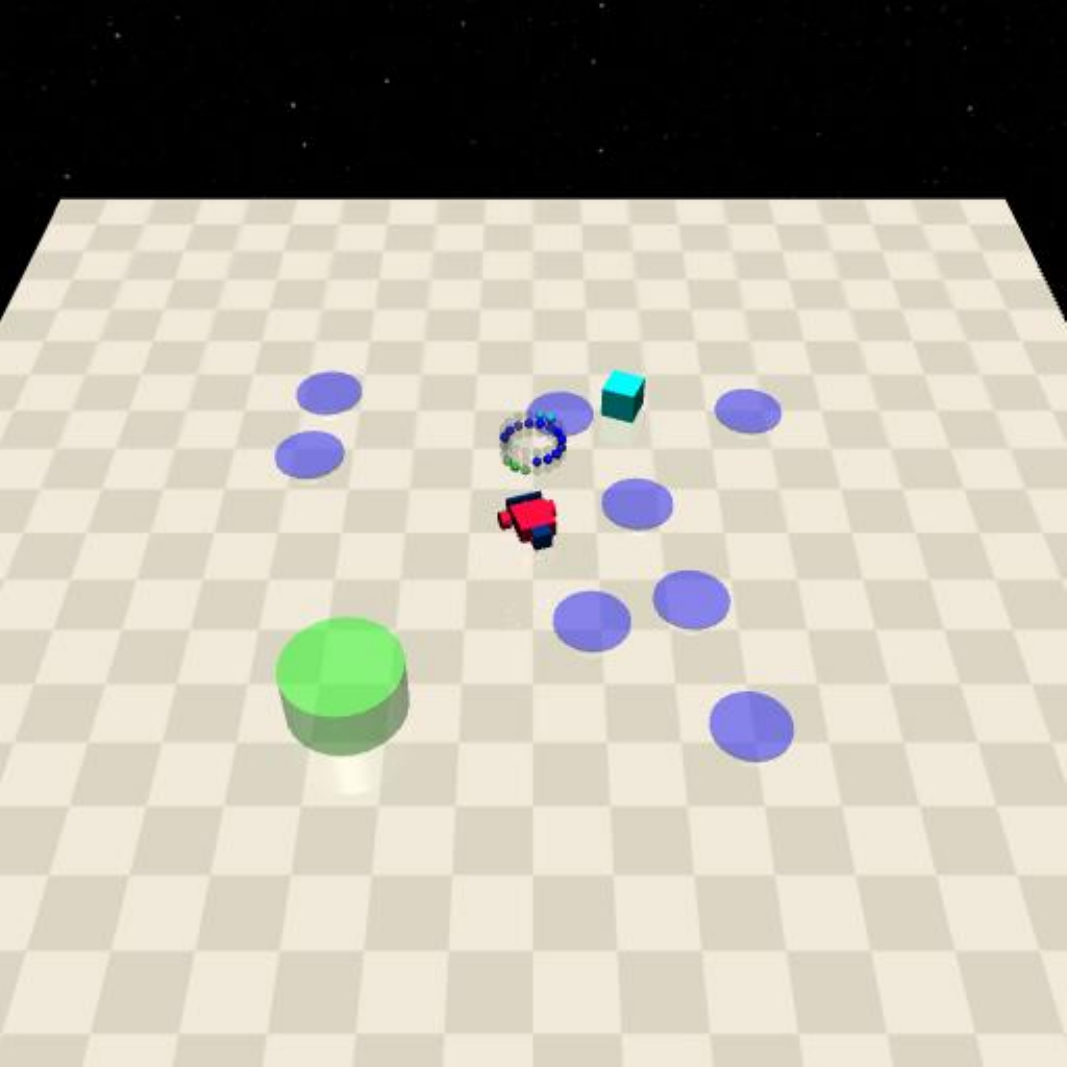} \\
  };

\path
  let \p2 = (m-1-3.south east) in coordinate (topRight) at (\x2,\y2);

\path
  let \p1 = (m-1-1.north west) in coordinate (topLeft) at (\x1,\y1);
\path
  let \p3 = (m-2-1.north west) in coordinate (botLeft) at (\x3,\y3);

\path
  let \p4 = (m-2-3.south east) in coordinate (botRight) at (\x4,\y4);

\begin{pgfonlayer}{background}
  \node (topbg) [fit=(topLeft) (topRight),
                 fill=ao!12, inner sep=0, rounded corners=2pt] {};
  \node (botbg) [fit=(botLeft) (botRight),
                 fill=bittersweet!12,  inner sep=0, rounded corners=2pt] {};
\end{pgfonlayer}

  \coordinate (YL) at ($(topbg.south)!0.5!(botbg.north)$);

  \coordinate (TopMidY) at ($(topbg.north)!0.5!(topbg.south)$);
  \coordinate (BotMidY) at ($(botbg.north)!0.5!(botbg.south)$);

\node[anchor=north, inner sep=0pt,
      font=\bfseries\fontsize{9}{9.6}\selectfont, text=ao!60!black]
     at ([yshift=-2mm]topbg.south)
     {(a) Training Environments};
\node[anchor=north, inner sep=0pt,
      font=\bfseries\fontsize{9}{9.6}\selectfont, text=bittersweet!60!black]
     at ([yshift=-2mm]botbg.south)
     {(b) Evaluation Environments};
\tikzset{colname/.style={font=\bfseries\small, inner sep=0pt}}

\coordinate (C1) at (m-1-1.center);
\coordinate (C2) at (m-1-2.center);
\coordinate (C3) at (m-1-3.center);

\node[colname, anchor=south] at ([yshift=1mm] topbg.north -| C1) {DarkRoom};
\node[colname, anchor=south] at ([yshift=1mm] topbg.north -| C2) {PointGoal};
\node[colname, anchor=south] at ([yshift=1mm] topbg.north -| C3) {CarGoal};

\end{tikzpicture}
\caption{\label{fig: env} Example training and evaluation layouts, adapted
from \citet{moeini2026safeincontextreinforcementlearning}. Training layouts are
center-oriented, whereas evaluation layouts are edge-oriented. We use this
shift for the DarkRoom, PointGoal and CarGoal test tasks.}
\end{figure*}

In each environment the tasks are split into training and test
sets. Following
\citet{moeini2026safeincontextreinforcementlearning}, layouts are sampled with
density $\propto\exp(\alpha\,d(x,c))$, where $d(x,c)$ is the distance from the
center of the grid or arena. A negative $\alpha$ favors the center and a
positive $\alpha$ the edge.
\begin{itemize}
\item \textbf{DarkRoom} is a partially observed $13\times13$ grid with one
hidden goal and 50 hidden obstacles, and the agent starts at the center.
Training uses 103 goals with three obstacle maps each, with goals and obstacles drawn
with $\alpha=-0.5$ (center oriented). The test set has 32 other goals, none of them a training
goal, each paired with a new obstacle map. Test goals and obstacles are drawn from edge-oriented distributions.
\item \textbf{PointGoal} and \textbf{CarGoal} are continuous navigation tasks
with a broken lidar, so the agent cannot sense goals or hazards until it reaches or
touches them. Goals and hazards are both placed with this density. Training
uses 128 center-oriented layouts with $\alpha=-0.2$. The test set has 30
edge-oriented layouts with $\alpha=+0.2$, which put goals and hazards where
they are rare in training.
\item In \textbf{HalfCheetah} and \textbf{Ant} the hidden velocity limit is
the default velocity threshold of Safety-Gymnasium times a factor in
$[0.5,2.0]$~\citep{ji2023safety}. Training uses 64 factors and the test
set 30 other factors.
\end{itemize}
Figure~\ref{fig: env} shows example training and test layouts.

\section{Teacher Data Details}\label{apx:teacher_data}
Table~\ref{tab:teacher} summarizes the environment-specific parts of the teacher
of Section~\ref{sec:teacher} and the size of the collected data. This appendix
gives the details.

\paragraph{Shared setup.} The budget range
$[B_{\min},B_{\max}]$ is $[1,15]$ in DarkRoom, $[10,50]$ in PointGoal and
CarGoal, and $[0,15]$ in the velocity tasks. We write $B=B_t^k$ for the
remaining budget of Eq.~\eqref{eq:collection}, which restarts at $B_0$ in every
episode. The teacher remembers, across the $K$ in-context episodes, where it has
been, where it paid cost and, once found, the task. Search explores more when
more budget remains. The expert uses the best behavior it knows whose cost fits
$B$. Histories of the same task and budget differ by the teacher's random seed,
which sets the Gaussian action noise (standard deviation 0.1 in the navigation
tasks and 0.05 in the velocity tasks), the tie-breaking order of the sweep, the
first speed and the speed step, which varies by up to 10\%.

\paragraph{DarkRoom.}
\begin{tightitems}
\item \emph{Switching event:} the first reward. Its position is the goal.
\item \emph{Search:} at each cell, the teacher tries a move it has not tried yet
with probability $0.1+0.9\min(B/B_{\max},1)$, and otherwise waits. After two
waits in a row it must try a move. A move that incurred cost is never tried
again. When no untried move is left at its cell, it walks the shortest known
path to the nearest cell that has one.
\item \emph{Expert:} it follows the shortest known path to the goal, which has
no observed cost, and then stays on the goal.
\end{tightitems}

\paragraph{PointGoal and CarGoal.}
\begin{tightitems}
\item \emph{Switching event:} the first reward. Its position gives the goal.
\item \emph{Planner:} a shortest-path planner on a 0.1\,m grid with
8-connected moves. A path costs its length plus a penalty for each cell it
enters within 0.25\,m of a position where cost was observed. The penalty is
1000 while $B<B_0/2$ and 0.3 otherwise. Ground with no observed cost has no
penalty, so the teacher avoids a hazard only after it has paid cost there in
the same $K$ in-context episodes. With no budget left, the robot only leaves
ground with known cost and then stands still.
\item \emph{Search:} the planner drives to the nearest unvisited point of a
0.5\,m grid over the arena. While $B<B_0/2$, it skips grid points within
0.35\,m of observed cost.
\item \emph{Expert:} the planner drives to the estimated goal, the mean position
of the rewarded steps.
\item \emph{Control:} a PPO waypoint controller drives toward a point 0.6\,m
ahead on the planned path.
\end{tightitems}

\paragraph{HalfCheetah and Ant.}
\begin{tightitems}
\item \emph{Switching event:} the first cost, which means the robot ran faster
than the hidden threshold. This only bounds the threshold, and the expert keeps
narrowing it.
\item \emph{Search:} the first speed is drawn per history from
$[0.8,1.2]$\,m/s (HalfCheetah) or $[0.5,0.8]$\,m/s (Ant). Every 25 steps the
teacher raises the commanded speed by about 0.5\,m/s (HalfCheetah) or
0.35\,m/s (Ant), as long as $B$ covers another probe. At the first cost it
returns to the last cost-free speed.
\item \emph{Expert:} it commands the slowest speed at which it has observed
cost, as long as $B$ covers that speed's cost per step, and otherwise the
fastest speed it has held without cost. When enough budget remains at the start
of an episode, it first spends 25 steps at the midpoint between these two
speeds, which narrows the gap between them.
\item \emph{Rules for both phases:} each episode starts with 25 steps at the
fastest cost-free speed found so far. A probe needs a reserve of at least 1
plus a running estimate of the cost paid while slowing down after a step down
in speed (initially 1). Cost in the first 10 steps after a change of command is
not attributed to the new command. If cost appears at a speed believed to be
safe, that speed is lowered by half a step.
\item \emph{Control:} a PPO speed controller tracks the commanded speed.
\end{tightitems}

\paragraph{PPO controllers.} Each controller is a PPO policy with two hidden
layers of 128 units. Training takes 6M, 8M, 20M and 30M steps in PointGoal,
CarGoal, HalfCheetah and Ant, with a checkpoint every 10\% of training. The
environment's own reward and cost are not used. The waypoint controllers of
PointGoal and CarGoal see the robot's accelerometer, velocimeter, gyroscope and
magnetometer (CarGoal also its rear wheel sensors) and the target's position
relative to the robot. They are trained on targets drawn uniformly over the
arena at least 0.6\,m away, and are rewarded for progress toward the target with
a bonus for reaching it. The speed controllers of HalfCheetah and Ant see the
environment observation and the target speed. They are trained on target
speeds in $[0.5,7]$\,m/s and $[0.25,5.5]$\,m/s and rewarded for tracking them.

\paragraph{Quality levels.} The $q=1$ controllers are the checkpoints with the
best validation score, saved at 10\%, 30\%, 100\% and 80\% of training in
PointGoal, CarGoal, HalfCheetah and Ant. For $q\approx0.5$, we measured $q$
(Eq.~\eqref{eq:quality}) on calibration tasks outside the training set, for
candidate checkpoints or, in DarkRoom, candidate random-action probabilities. We took
the candidate closest to 0.5 within $[0.4,0.6]$. Table~\ref{tab:quality} gives the chosen $q$ and
how often the teacher goes over its budget in the dynamic-budget histories.
\begin{table}[h]
\centering
\caption{Chosen quality of the weaker expert and the share of teacher episodes over budget in the dynamic-budget histories.}
\label{tab:quality}
\small
\begin{tabular}{lccc}
\toprule
 & Chosen $q$ & \multicolumn{2}{c}{Episodes over budget} \\
\cmidrule(lr){3-4}
Environment & ($q\approx0.5$ level) & $q\approx0.5$ & $q=1$ \\
\midrule
DarkRoom (random-action probability 0.3) & 0.49 & 5.3\% & 5.0\% \\
PointGoal & 0.55 & 25.8\% & 11.5\% \\
CarGoal & 0.43 & 33.4\% & 12.7\% \\
HalfCheetah & 0.44 & 25.6\% & 12.3\% \\
Ant & 0.51 & 12.5\% & 7.3\% \\
\bottomrule
\end{tabular}
\end{table}
With the $q\approx0.5$ expert, the teacher also has a higher cost in every
environment: three to four times as high in PointGoal and CarGoal, and only
slightly higher in Ant (2.83 against 2.80 per in-context episode). It finds the
task less often in PointGoal, HalfCheetah and Ant, in 66\%, 73\% and 54\% of
the histories, against 87\% to 98\% with $q=1$.

\paragraph{Training mixture.} We collect 16 histories per task, budget regime
and quality level. The main models train on the dynamic-budget histories of both
quality levels, half of the histories from each. Counted in steps, the split is
also even in DarkRoom and the velocity tasks. In PointGoal and CarGoal the
$q\approx0.5$ episodes are longer (about 56 against 23 steps), so the $q=1$
histories supply about 29\% of the training steps. The expert-share models of
Appendix~\ref{apx:mix} use 16 dynamic-budget histories per task, of which 0, 4,
8, 12 or 16 come from the $q=1$ expert and the rest from the $q\approx0.5$
expert.

\paragraph{What the teacher sees.} To check that the teacher uses only what the
student sees, we run it again on the logged observations alone. In DarkRoom this
reproduces every history. In the other environments we ran this check on
separate histories collected the same way, not on the training histories.

\begin{table}[h]
\centering
\caption{The teacher and its data. Top: the environment-specific parts of the
teacher. Everything else is shared by all five environments. Bottom: teacher
data per environment, over both budget regimes and both quality levels.}
\label{tab:teacher}
\small
\begin{tabular}{p{2.1cm}p{2.3cm}p{4.4cm}p{4.6cm}}
\toprule
Environment & Switching event & Search & Expert of quality $q$ \\
\midrule
DarkRoom & first reward, at the goal & tries new moves, more often when more
budget is left. Never repeats a move that cost & takes the shortest known path
without cost. For $q<1$, mixes in random actions at a fixed rate \\
PointGoal, CarGoal & first reward, at the goal & drives through a grid of points
over the arena. Goes less far around known hazards when more budget is left &
drives to the goal position estimated from the history, using a PPO controller.
For $q<1$, the controller is saved earlier in training \\
HalfCheetah, Ant & first cost, from running too fast & runs faster every 25
steps while the budget allows another try & runs at the fastest speed whose
known cost fits the budget, using a PPO controller. For $q<1$, the controller is
saved earlier in training \\
\bottomrule
\end{tabular}

\medskip
\begin{tabular}{lrrrr}
\toprule
Environment & Training tasks & $K$ & Histories & Steps \\
\midrule
DarkRoom & 309 (103 goals $\times$ 3 maps) & 50 & 19{,}776 & 29.7M \\
PointGoal & 128 layouts & 20 & 8{,}192 & 6.5M \\
CarGoal & 128 layouts & 20 & 8{,}192 & 6.3M \\
HalfCheetahVelocity & 64 thresholds & 10 & 4{,}096 & 8.2M \\
AntVelocity & 64 thresholds & 10 & 4{,}096 & 8.2M \\
\bottomrule
\end{tabular}
\end{table}

\section{Training Details}\label{apx: training details}
\paragraph{Baselines.}
The sequence models share the backbone in Table~\ref{tab: sp parameters} and the same tokenization, four tokens per step
for state, action, reward, and cost. Safe AD and A-Lag add both RTG and CTG embeddings to the state token at every step. A-Lag also has the cost head,
which changes actions only through the deployment-time multiplier. In our CDT implementation, RTG and CTG are carried by context-level prefix tokens rather than per step. This differs from the original CDT, which conditions on return and cost at every step, and CDT uses the same training settings as Safe AD and A-Lag rather than separately tuned ones. The fixed-budget models of RQ1 differ from the main models only in the training data. The two gradient-based safe
meta-RL baselines, MAML with a cost penalty and SafeMeta, adapt policy parameters on each evaluation task and therefore share none of this backbone.
They keep the settings of the preliminary study (bottom of
Table~\ref{tab: sp parameters}).

\paragraph{A-Lag implementation.} The cost head is one linear layer on the
state-position representation $\mathbf u_p$. For discrete actions it outputs one
score per action. For continuous actions it outputs $b$, $g_d$ and a raw
curvature per action dimension, with $\kappa_d$ the softplus of the raw value.
At initialization $b=g_d=0$ and $\kappa_d=10^{-3}$, so the reweighted policy
starts close to the base policy. The loss weight is $\eta_C=1$ in every
environment. The reweighted Gaussian of Section~\ref{sec:alag_head} is the
distribution of the sampled action before any clipping to the action range,
and Proposition~\ref{prop:alag_mono} concerns that distribution. Within an
episode the RTG and CTG inputs fall by the received reward and the incurred
cost, the CTG not below zero, and both restart at their deployment values in
the next episode.

\paragraph{Hyperparameter Details.}
Table~\ref{tab: sp parameters} reports the configurations used in our experiments. All sequence models are trained with AdamW
($\beta=(0.9,0.99)$, weight decay $0.01$, gradient-norm clipping at $1.0$) under a cosine schedule with warmup over the first $2.5\%$ (DarkRoom) or $5\%$
(all other environments) of the update budget, for 100{,}000 updates with seeds 42, 43 and 44, and we evaluate the final checkpoint. CDT trains on the
same dynamic-budget histories as Safe AD and A-Lag. PointGoal and CarGoal share the navigation settings below. HalfCheetahVelocity and AntVelocity share the velocity settings.
\begin{table*}[t]
\centering
\caption{Training settings. Top: the sequence models (Safe AD, A-Lag and CDT).
Num Layers is $L$ in Section~\ref{sec:causal_sequence}. Seq Len counts environment steps, not token positions. Safe AD and A-Lag use four positions per step plus two constant prefix positions. Token Embedding Dim
is the width of the per-modality encoders before they are projected to the
Hidden Dim. The navigation tasks use four gradient
accumulation steps, giving an effective batch size of 32. Action-input dropout
sets each component of an action input in the history to zero with the given
probability during training. The action targets are unchanged. Bottom: the gradient-based safe meta-RL
baselines, SafeMeta and MAML with a cost penalty, as in the preliminary study.
The navigation tasks are PointGoal and CarGoal, and the velocity tasks are HalfCheetah and Ant.}
\label{tab: sp parameters}
\small
\begin{tabular}{@{}llll@{}}
\toprule
Parameter & DarkRoom & PointGoal/CarGoal & Velocity tasks \\
\midrule
Hidden Dim & 256 & 256 & 256 \\
Token Embedding Dim & 64 & 128 & 128 \\
Num Layers & 6 & 6 & 6 \\
Num Heads & 4 & 8 & 8 \\
Seq Len & 100 & 500 & 400 \\
Attention Dropout & 0.5 & 0.5 & 0.1 \\
Residual Dropout & 0.3 & 0.3 & 0.1 \\
Embedding Dropout & 0.2 & 0.2 & 0.1 \\
Action-Input Dropout & 0 & 0.1 & 0.1 \\
Learning Rate & $5\times10^{-5}$ & $3\times10^{-4}$ & $3\times10^{-4}$ \\
Batch Size & 256 & 8 & 32 \\
Gradient Accumulation & 1 & 4 & 1 \\
Update Budget & 100,000 & 100,000 & 100,000 \\
In-Context Episodes $K$ & 50 & 20 & 10 \\
Episode Time Limit & 30 & 75 & 200 \\
\bottomrule
\end{tabular}

\medskip
\begin{tabular}{@{}lll@{}}
\toprule
Safe meta-RL baselines & DarkRoom & Navigation and velocity tasks \\
\midrule
Hidden layers & (64, 64) & (64, 512, 64) \\
Min/Max batch size & 500/1500 & 1500/1500 \\
Policy learning rate & 1e-3 & 1e-3 \\
Value/Cost learning rate & 3e-2/1e-1 & 3e-2/1e-1 \\
Discount factor $\gamma$ & 0.99 & 0.99 \\
GAE parameter & 0.95 & 0.95 \\
Max KL divergence & 1e-3 & 1e-3 \\
Lagrangian weight & 1.0 & 1.0 \\
Policy optimizer & Adam & Adam \\
Value/Cost optimizer & L-BFGS & L-BFGS \\
\bottomrule
\end{tabular}
\end{table*}

\section{Teacher-Data Coverage}\label{apx:coverage}
\begin{figure*}[t]
  \centering
  \includegraphics[width=\textwidth]{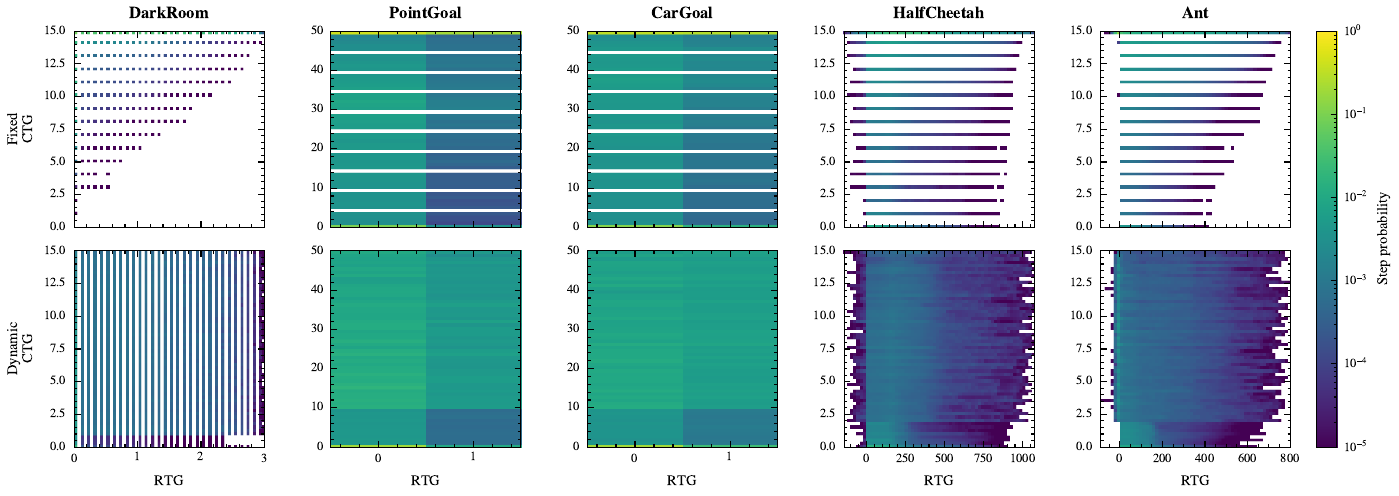}
  \caption{Occupancy of the (RTG, CTG) plane in the teacher data,
  fixed budgets (top) and dynamic budgets (bottom), both quality levels. Color
  is the share of steps on a logarithmic scale. In PointGoal and CarGoal the
  RTG is 0 or 1, so it is drawn as two columns. Under a fixed budget the
  remaining budget starts at $B_{\max}$ and falls in steps of the incurred cost,
  which leaves gaps between the CTG values that occur.}
  \label{fig:coverage}
\end{figure*}

We measure how much of the plane of the two per-step control inputs, RTG and
CTG, the fixed- and dynamic-budget histories cover. In each environment we
rescale each coordinate to $[0,1]$, mapping its 1st and 99th percentiles over
the fixed- and dynamic-budget data together to 0 and 1 and clipping values
outside, so that a few extreme values do not stretch the grid. We then divide
the plane into a $20\times20$ grid. We report
the number of occupied cells and, with $n_b$ steps in cell $b$, the normalized
occupancy entropy
\begin{equation}
p_b=\frac{n_b}{\sum_{b'}n_{b'}},\qquad
\mathrm{Entropy}=-\frac{\sum_{b:p_b>0}p_b\log p_b}{\log 400}.
\end{equation}
Table~\ref{tab:coverage} in Section~\ref{sec:rq1} gives the values, and
Figure~\ref{fig:coverage} shows the occupancy. We also measured, on the same
grid, a window coverage: the plane of return and cost summed over non-overlapping windows of
100 steps in DarkRoom, 500 in PointGoal and CarGoal and 1{,}000 in the velocity
tasks. There the dynamic-budget histories cover slightly less than the
fixed-budget ones in DarkRoom and the velocity tasks and about the same in
PointGoal and CarGoal.

\section{Deployment-Time Settings and Multiplier Trajectories}
\label{apx:controller}
\begin{figure*}[t]
  \centering
  \includegraphics[width=\textwidth]{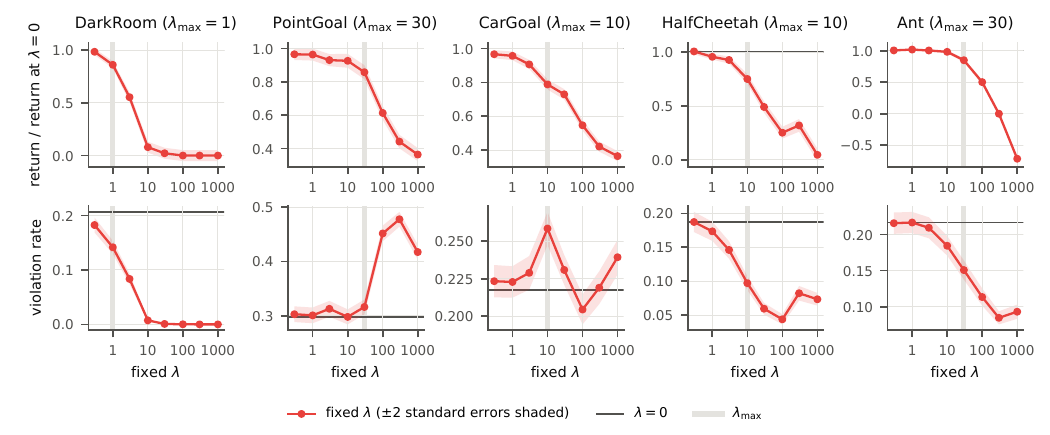}
  \caption{Fixed multipliers on the training tasks (the reported A-Lag models,
  averaged over three seeds). Top: final return relative to $\lambda=0$, with
  a band of $\pm2$ standard errors of the paired difference. Bottom:
  violation rate with a band of $\pm2$ standard errors. The wide gray line marks $\lambda_{\max}$. Larger fixed
  multipliers lower the return in every environment, and the violation rate in
  DarkRoom, HalfCheetah and Ant. The violation rate is nonmonotonic in PointGoal and CarGoal and can increase as the multiplier grows, including within the selected range in CarGoal. A-Lag uses
  this sweep only to set $\lambda_{\max}$ and tunes $\lambda$ within
  $[0,\lambda_{\max}]$ during the test deployment, from the costs it incurs.}
  \label{fig:grid}
\end{figure*}

The upper bound $\lambda_{\max}$ in Eq.~\eqref{eq:dual_update} is fixed per environment on a fixed
subset of the training tasks and with the trained A-Lag models, by a fixed rule. A sweep of fixed
multipliers $\lambda\in\{0.3,1,3,10,30,100,300,1000\}$ sets the upper bound:
$\lambda_{\max}$ is the largest tested value up to which the mean final return
stays above 75\% of that at $\lambda=0$. This gives
$\lambda_{\max}=1$ in DarkRoom, 30 in PointGoal, 10 in CarGoal, 10 in
HalfCheetah and 30 in Ant (Figure~\ref{fig:grid}). The fixed multiplier of
Table~\ref{tab:fixed} comes from the same sweep, by a second fixed rule: among the values up to $\lambda_{\max}$ whose return is not
clearly below that at $\lambda=0$ (by its $\pm2$ standard-error band), we take the one with the lowest violation
rate. This gives 0.3 in DarkRoom, 1 in PointGoal, 0.3 in HalfCheetah and 10 in
Ant. In CarGoal no tested value met the rule, so we use the one whose return came closest, 0.3.

In every environment the multiplier update uses the step size $\eta=\lambda_{\max}$, the bounds $[0,\lambda_{\max}]$ and the scale $B_{\max}$,
the upper end of the budget range (15 in DarkRoom, 50 in PointGoal and
CarGoal, 15 in the velocity tasks). We chose the step size once in DarkRoom, from
$\{1,3,10\}\cdot\lambda_{\max}$, and use it
in every environment.

\begin{figure*}[t]
  \centering
  \includegraphics[width=\textwidth]{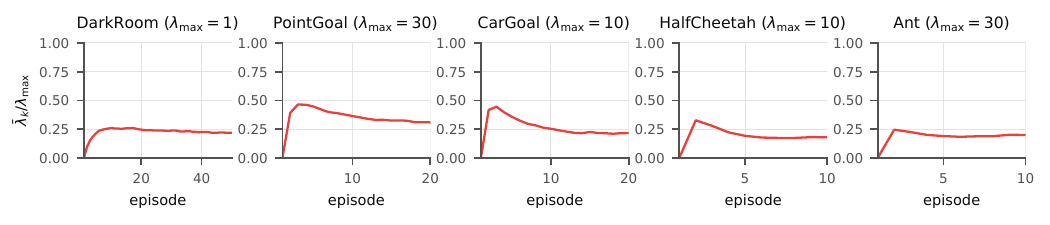}
  \caption{A-Lag's mean multiplier per episode on the training tasks,
  as a fraction of $\lambda_{\max}$, averaged over three seeds.}
  \label{fig:lambda}
\end{figure*}

Figure~\ref{fig:lambda} shows A-Lag's mean multiplier in each episode
on the training tasks, averaged over three seeds. On average the multiplier peaks at 0.25 to 0.46 of $\lambda_{\max}$, within the first three
episodes in the navigation and velocity tasks and around the tenth episode in
DarkRoom, and then declines slowly.

\section{Proofs of the A-Lag Properties}\label{apx:proofs}
This appendix proves the statements of Section~\ref{sec:alag_theory}. The
context is fixed throughout, so $\mathbf u=\mathbf u_p$ does not change. We
write $\hat c(a)$ for $\hat c(\mathbf u,a)$ and $\pi_\theta(a)$ for $\pi_\theta(a\mid\mathbf u)$,
so that $\pi_\lambda(a)=\pi_\theta(a)\,e^{-\lambda \hat c(a)}/Z_\lambda$ with
$Z_\lambda=\mathbb E_{\pi_\theta}[e^{-\lambda \hat c}]$. The last paragraph shows
that every expectation used below is finite.

\paragraph{KL characterization.} Since
$\log\pi_\lambda(a)=\log\pi_\theta(a)-\lambda \hat c(a)-\log Z_\lambda$, every
action distribution $\pi$ with finite objective satisfies
\begin{equation*}
\mathrm{KL}(\pi\,\|\,\pi_\theta)+\lambda\,\mathbb E_{\pi}[\hat c]
=\mathbb E_{A\sim\pi}\Bigl[\log\frac{\pi(A)}{\pi_\theta(A)}+\lambda \hat c(A)\Bigr]
=\mathbb E_{A\sim\pi}\Bigl[\log\frac{\pi(A)}{\pi_\lambda(A)}\Bigr]-\log Z_\lambda
=\mathrm{KL}(\pi\,\|\,\pi_\lambda)-\log Z_\lambda .
\end{equation*}
The term $\log Z_\lambda$ does not depend on $\pi$, and
$\mathrm{KL}(\pi\,\|\,\pi_\lambda)\ge0$ with equality only for
$\pi=\pi_\lambda$. Hence $\pi_\lambda$ is the unique minimizer.

\paragraph{Proof of Proposition~\ref{prop:alag_mono}.} Let
$N_\lambda=\mathbb E_{\pi_\theta}[\hat c\,e^{-\lambda \hat c}]$ and
$M_\lambda=\mathbb E_{\pi_\theta}[\hat c^2e^{-\lambda \hat c}]$. Then
$\mathbb E_{\pi_\lambda}[\hat c]=N_\lambda/Z_\lambda$ and
$\mathbb E_{\pi_\lambda}[\hat c^2]=M_\lambda/Z_\lambda$. Differentiating under the
expectation, which the next paragraph justifies, gives
$Z_\lambda'=-N_\lambda$ and $N_\lambda'=-M_\lambda$. Taking the derivative of $N_\lambda/Z_\lambda$ gives
\begin{equation*}
\frac{d}{d\lambda}\,\mathbb E_{\pi_\lambda}[\hat c]
=\frac{N_\lambda'Z_\lambda-N_\lambda Z_\lambda'}{Z_\lambda^2}
=-\frac{M_\lambda}{Z_\lambda}+\Bigl(\frac{N_\lambda}{Z_\lambda}\Bigr)^2
=-\mathbb E_{\pi_\lambda}[\hat c^2]+\bigl(\mathbb E_{\pi_\lambda}[\hat c]\bigr)^2
=-\operatorname{Var}_{\pi_\lambda}(\hat c)\le0 .
\end{equation*}

\paragraph{Exchanging derivative and expectation.} For finite action sets every
expectation is a finite sum, and the exchange is immediate. In the Gaussian
case $\pi_\theta$ is a diagonal Gaussian and
$\hat c(a)=b+\sum_d\bigl(g_da_d+\tfrac12\kappa_da_d^2\bigr)$ with $\kappa_d\ge0$
(Section~\ref{sec:alag_head}). Fix $\Lambda>0$. For
$\lambda\in[0,\Lambda]$ we have $e^{-\lambda \hat c(a)}\le1+e^{-\Lambda \hat c(a)}$,
because the left side is at most $1$ when $\hat c(a)\ge0$ and at most
$e^{-\Lambda \hat c(a)}$ when $\hat c(a)<0$. Since $\kappa_d\ge0$,
$e^{-\Lambda \hat c(a)}\le e^{-\Lambda b-\Lambda\sum_dg_da_d}$. For
$k\in\{0,1,2\}$, $|\hat c(a)|^ke^{-\lambda \hat c(a)}$ is therefore bounded, uniformly
in $\lambda\in[0,\Lambda]$, by a polynomial in $a$ times one plus the
exponential of a linear function of $a$. This bound is integrable under a
Gaussian. So $Z_\lambda$, $N_\lambda$ and $M_\lambda$ are finite, and because the
bound does not depend on $\lambda$, we can differentiate $Z_\lambda$ and
$N_\lambda$ under the expectation on $[0,\Lambda)$. Because $\Lambda$ is arbitrary, this holds for
every $\lambda\ge0$.

\section{Expert Quality}\label{apx:mix}
\begin{figure*}[t]
  \centering
  \includegraphics[width=\textwidth]{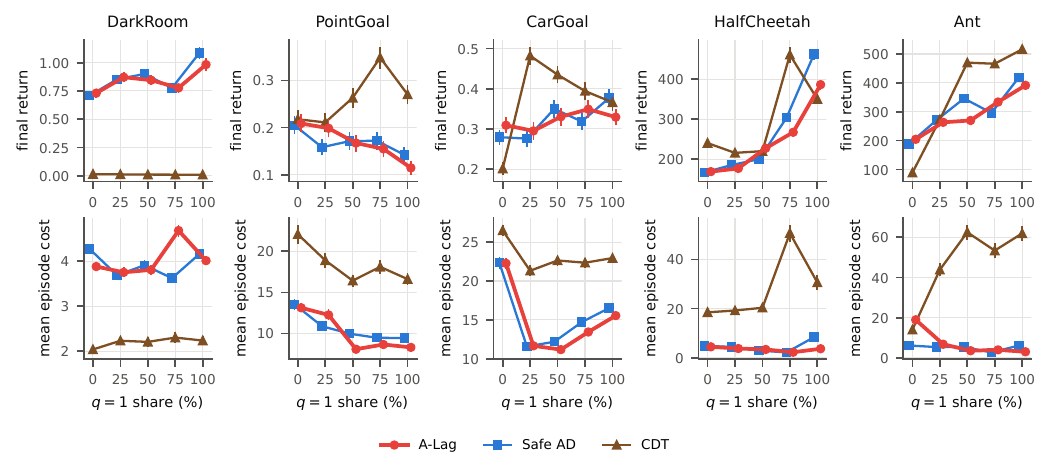}
  \caption{Final return and mean episode cost
  on the test tasks against the share of $q=1$ histories (16 histories
  per task), averaged over three seeds, with bands of $\pm2$ standard errors. The main models and the violation rates are
  in Table~\ref{tab:mix}.}
  \label{fig:mix_share}
\end{figure*}

We train each method on 16 dynamic-budget histories per task, of which 0, 25,
50, 75 or 100\% come from the teacher with the $q=1$ expert and the rest
with the $q\approx0.5$ expert. The main models use 32 histories per task at 50\%, so
they differ from the 50\% runs in data size as well. Figure~\ref{fig:mix_share}
shows the final return and the mean episode cost on the test tasks, and
Table~\ref{tab:mix} gives the numbers, including the violation rates.

In DarkRoom and the velocity tasks, every method that learns the task reaches
its highest return at 75 or 100\% $q=1$ histories and its lowest at 0\%, except
CDT in HalfCheetah, whose lowest is at 25\%. Safe AD, for example, returns 1.087
at 100\% against 0.709 at 0\% in DarkRoom and 462.5 against 166.4 in
HalfCheetah. The cost does not follow the return. Safe AD at 100\% costs more
than at 75\% in HalfCheetah and Ant, and it also violates the budget more often
there. In the velocity tasks A-Lag's cost at 100\% is lower than at 0\%.
CDT's cost grows with its return in the velocity tasks. PointGoal reverses the
trend for Safe AD and A-Lag. Their return is highest at 0\% and lowest at
100\%, while CDT's is highest at 75\%. In CarGoal, Safe AD and A-Lag return less
at every share than on the main data, which has twice as many histories
per task, while CDT at 25\% matches its main model.

\begin{table}[t]
\centering
\caption{Expert-share study on the test split: final return, mean episode cost and violation rate of models trained on 16 dynamic-budget histories per task with the given share of the $q=1$ expert (the rest $q\approx0.5$), and of the main models (50\%, 32 histories per task). Means over three seeds. Bold marks the best value in each row: the highest return and the lowest cost and violation rate.}
\label{tab:mix}
\scriptsize
\setlength{\tabcolsep}{3pt}
\par\textit{Final return}\par\smallskip
\begin{adjustbox}{max width=\textwidth}
\begin{tabular}{@{}llrrrrrr@{}}
\toprule
Environment & Method & 0\% & 25\% & 50\% & 75\% & 100\% & Main \\
\midrule
DarkRoom & Safe AD & $0.709$ & $0.853$ & $0.900$ & $0.777$ & $\mathbf{1.087}$ & $0.894$ \\
 & CDT & $\mathbf{0.014}$ & $0.012$ & $0.010$ & $0.009$ & $0.009$ & $0.010$ \\
 & A-Lag & $0.731$ & $0.872$ & $0.846$ & $0.777$ & $\mathbf{0.984}$ & $0.832$ \\
\addlinespace
PointGoal & Safe AD & $\mathbf{0.205}$ & $0.159$ & $0.171$ & $0.172$ & $0.142$ & $0.173$ \\
 & CDT & $0.218$ & $0.211$ & $0.263$ & $\mathbf{0.348}$ & $0.270$ & $0.326$ \\
 & A-Lag & $\mathbf{0.209}$ & $0.199$ & $0.167$ & $0.155$ & $0.115$ & $0.164$ \\
\addlinespace
CarGoal & Safe AD & $0.280$ & $0.276$ & $0.350$ & $0.319$ & $0.377$ & $\mathbf{0.489}$ \\
 & CDT & $0.201$ & $\mathbf{0.482}$ & $0.435$ & $0.394$ & $0.366$ & $0.468$ \\
 & A-Lag & $0.309$ & $0.295$ & $0.330$ & $0.349$ & $0.329$ & $\mathbf{0.461}$ \\
\addlinespace
HalfCheetah & Safe AD & $166.4$ & $185.1$ & $199.9$ & $303.5$ & $\mathbf{462.5}$ & $206.7$ \\
 & CDT & $240.0$ & $215.5$ & $219.8$ & $\mathbf{460.8}$ & $349.9$ & $284.9$ \\
 & A-Lag & $168.6$ & $176.6$ & $227.1$ & $267.0$ & $\mathbf{386.1}$ & $218.5$ \\
\addlinespace
Ant & Safe AD & $188.7$ & $270.9$ & $346.0$ & $295.8$ & $\mathbf{419.2}$ & $350.9$ \\
 & CDT & $89.0$ & $273.0$ & $470.0$ & $466.1$ & $\mathbf{516.2}$ & $387.0$ \\
 & A-Lag & $204.6$ & $263.4$ & $269.8$ & $333.6$ & $\mathbf{391.4}$ & $286.4$ \\
\bottomrule
\end{tabular}
\end{adjustbox}
\par\medskip\textit{Mean episode cost}\par\smallskip
\begin{adjustbox}{max width=\textwidth}
\begin{tabular}{@{}llrrrrrr@{}}
\toprule
Environment & Method & 0\% & 25\% & 50\% & 75\% & 100\% & Main \\
\midrule
DarkRoom & Safe AD & $4.27$ & $3.69$ & $3.91$ & $\mathbf{3.62}$ & $4.16$ & $3.85$ \\
 & CDT & $\mathbf{2.04}$ & $2.23$ & $2.21$ & $2.30$ & $2.24$ & $2.14$ \\
 & A-Lag & $3.88$ & $3.75$ & $3.81$ & $4.68$ & $4.01$ & $\mathbf{3.45}$ \\
\addlinespace
PointGoal & Safe AD & $13.54$ & $10.89$ & $9.97$ & $9.48$ & $\mathbf{9.44}$ & $10.04$ \\
 & CDT & $22.07$ & $18.90$ & $\mathbf{16.43}$ & $18.11$ & $16.61$ & $16.98$ \\
 & A-Lag & $13.13$ & $12.25$ & $\mathbf{8.08}$ & $8.66$ & $8.31$ & $9.69$ \\
\addlinespace
CarGoal & Safe AD & $22.34$ & $\mathbf{11.58}$ & $12.19$ & $14.71$ & $16.53$ & $14.77$ \\
 & CDT & $26.49$ & $21.32$ & $22.64$ & $22.35$ & $22.94$ & $\mathbf{20.97}$ \\
 & A-Lag & $22.27$ & $11.66$ & $\mathbf{11.18}$ & $13.44$ & $15.55$ & $11.80$ \\
\addlinespace
HalfCheetah & Safe AD & $4.90$ & $4.45$ & $2.96$ & $\mathbf{2.23}$ & $8.41$ & $3.70$ \\
 & CDT & $\mathbf{18.50}$ & $19.24$ & $20.35$ & $50.65$ & $30.63$ & $22.06$ \\
 & A-Lag & $4.48$ & $3.80$ & $3.41$ & $\mathbf{2.29}$ & $3.71$ & $3.59$ \\
\addlinespace
Ant & Safe AD & $6.15$ & $5.38$ & $5.38$ & $\mathbf{2.72}$ & $6.15$ & $5.55$ \\
 & CDT & $\mathbf{14.06}$ & $43.86$ & $62.53$ & $53.41$ & $61.97$ & $43.24$ \\
 & A-Lag & $18.95$ & $6.80$ & $3.55$ & $3.98$ & $\mathbf{3.02}$ & $4.08$ \\
\bottomrule
\end{tabular}
\end{adjustbox}
\par\medskip\textit{Violation rate}\par\smallskip
\begin{adjustbox}{max width=\textwidth}
\begin{tabular}{@{}llrrrrrr@{}}
\toprule
Environment & Method & 0\% & 25\% & 50\% & 75\% & 100\% & Main \\
\midrule
DarkRoom & Safe AD & $0.250$ & $0.211$ & $0.227$ & $\mathbf{0.200}$ & $0.239$ & $0.229$ \\
 & CDT & $\mathbf{0.092}$ & $0.115$ & $0.111$ & $0.116$ & $0.108$ & $0.108$ \\
 & A-Lag & $0.213$ & $0.206$ & $0.206$ & $0.274$ & $0.214$ & $\mathbf{0.180}$ \\
\addlinespace
PointGoal & Safe AD & $0.170$ & $0.146$ & $0.143$ & $\mathbf{0.133}$ & $0.140$ & $0.139$ \\
 & CDT & $0.222$ & $0.222$ & $0.202$ & $0.217$ & $0.204$ & $\mathbf{0.195}$ \\
 & A-Lag & $0.190$ & $0.184$ & $\mathbf{0.117}$ & $0.122$ & $0.126$ & $0.137$ \\
\addlinespace
CarGoal & Safe AD & $0.257$ & $\mathbf{0.159}$ & $0.171$ & $0.207$ & $0.240$ & $0.207$ \\
 & CDT & $0.293$ & $0.272$ & $0.288$ & $0.281$ & $0.299$ & $\mathbf{0.270}$ \\
 & A-Lag & $0.259$ & $0.162$ & $\mathbf{0.159}$ & $0.197$ & $0.226$ & $0.164$ \\
\addlinespace
HalfCheetah & Safe AD & $0.173$ & $0.161$ & $0.112$ & $\mathbf{0.105}$ & $0.298$ & $0.123$ \\
 & CDT & $0.338$ & $0.282$ & $0.235$ & $0.363$ & $\mathbf{0.212}$ & $0.239$ \\
 & A-Lag & $0.169$ & $0.148$ & $0.155$ & $\mathbf{0.108}$ & $0.162$ & $0.158$ \\
\addlinespace
Ant & Safe AD & $0.272$ & $0.306$ & $0.331$ & $\mathbf{0.154}$ & $0.480$ & $0.356$ \\
 & CDT & $\mathbf{0.211}$ & $0.358$ & $0.490$ & $0.406$ & $0.475$ & $0.364$ \\
 & A-Lag & $0.341$ & $0.261$ & $\mathbf{0.164}$ & $0.173$ & $0.193$ & $0.184$ \\
\bottomrule
\end{tabular}
\end{adjustbox}
\end{table}

\section{Additional Results}\label{apx:tables}
Table~\ref{tab:main} gives the numbers behind RQ2 and RQ3.

\paragraph{Multiplier ablation.} We evaluate the same A-Lag model at $\lambda=0$. Only the multiplier differs, so this comparison measures its effect without retraining. The cost loss trains the cost head without updating the backbone, so A-Lag at $\lambda=0$ uses the same policy-training loss as Safe AD, and differences between the two can come from training variation (Table~\ref{tab:seed_alag}). The multiplier lowers mean cost in every environment but Ant. It lowers return in DarkRoom and CarGoal, by 0.044 and 0.045, and slightly in HalfCheetah and Ant. In Ant, A-Lag's lower cost than Safe AD is already present at $\lambda=0$, which also returns 61.9 less than Safe AD. Applying the multiplier raises Ant's cost. Table~\ref{tab:main} gives the full comparison.

\paragraph{Adaptive vs.\ fixed multiplier.} Table~\ref{tab:fixed} compares the
adaptive multiplier with a fixed one chosen on training tasks, on the same
weights. The two rules differ in penalty size as well as in feedback. On
training tasks the adaptive multiplier peaks at 0.25 to 0.46 of
$\lambda_{\max}$ (Figure~\ref{fig:lambda}), above the fixed value in PointGoal,
CarGoal and HalfCheetah, similar to it in DarkRoom and below it in Ant. The adaptive one lowers the mean violation rate in DarkRoom and HalfCheetah, by 0.010 and 0.022. Its mean return is about the same in DarkRoom and lower in HalfCheetah. In Ant it returns 12.0 more at a higher cost. In PointGoal it returns more at a lower cost, with the same violation rate. In HalfCheetah the fixed $\lambda=0.3$ performs almost like $\lambda=0$ (Table~\ref{tab:main}), so there the comparison is close to one against no multiplier. In CarGoal the adaptive multiplier lowers the mean violation rate by 0.023 and also lowers mean return. It also lowers the excess cost
in every environment but Ant.

\paragraph{RQ1 violation rates in DarkRoom.} In DarkRoom the dynamic-budget models return 0.167 (Safe AD) and 0.123 (A-Lag) more than the fixed-budget models
and violate the budget 0.085 and 0.097 less often.

\paragraph{Sensitivity to data composition.} The all-five-environment mean-cost comparison refers to the main 50\% expert mixture with 32 histories per task. In the expert-share study, which uses 16 histories per task, A-Lag has lower mean cost in 16 of the 25 combinations of environment and expert share and lower mean violation rate in 16 (Table~\ref{tab:mix}). At the 100\% share, A-Lag also has lower mean cost than Safe AD in all five environments, at a lower mean return in all five. At the 0\% strong-expert share in Ant, its mean cost is 18.95 against Safe AD's 6.15. The main result therefore does not extend to every data mixture.

\paragraph{Scope of the controls.} RQ1 compares training at $B_{\max}$ with training over dynamic budgets, so its effect includes training--test budget mismatch. The PPO-Lagrangian comparison uses a different protocol and does not isolate the teacher's effect (Appendix~\ref{apx:ppolag}). We do not include a budget-unaware teacher control or a teacher reference evaluated under the student test protocol.

\paragraph{Deployment-time controls.} Safe AD has no comparable deployment-time setting to tune. We have not evaluated a fixed multiplier matched to the adaptive rule's mean magnitude or a learned shield on the same backbone. The fixed-multiplier navigation sweeps do not show a monotone reduction in violations (Figure~\ref{fig:grid}).

\begin{table*}[t]
\centering
\caption{Test-split results of Safe AD, of A-Lag with the multiplier held at zero and of A-Lag, whose multiplier adapts after every episode. All columns are means over three seeds. Bold marks the best of the three methods in each row: the highest return and the lowest cost, violation rate and excess cost. Return is the final return, the mean over the last three episodes, and is the success rate in PointGoal and CarGoal. Cost is the mean episode cost. The violation rate is the fraction of episodes whose cost exceeds the deployment's budget $B_0$, so it counts how often a method goes over its budget, not by how much. Excess cost is the mean over episodes of $\max(0,G_c-B_0)$.}
\label{tab:main}
\footnotesize
\setlength{\tabcolsep}{3pt}
\begin{adjustbox}{max width=\textwidth}
\begin{tabular}{@{}llccc@{}}
\toprule
Environment & Metric & Safe AD & A-Lag, $\lambda=0$ & A-Lag \\
\midrule
DarkRoom & Return & $\mathbf{0.894}$ & $0.877$ & $0.832$ \\
 & Cost & $3.85$ & $3.64$ & $\mathbf{3.45}$ \\
 & Violation & $0.229$ & $0.209$ & $\mathbf{0.180}$ \\
 & Excess cost & $0.60$ & $0.54$ & $\mathbf{0.39}$ \\
\addlinespace
PointGoal & Return & $\mathbf{0.173}$ & $0.156$ & $0.164$ \\
 & Cost & $10.04$ & $10.20$ & $\mathbf{9.69}$ \\
 & Violation & $0.139$ & $0.139$ & $\mathbf{0.137}$ \\
 & Excess cost & $4.20$ & $4.24$ & $\mathbf{3.88}$ \\
\addlinespace
CarGoal & Return & $0.489$ & $\mathbf{0.506}$ & $0.461$ \\
 & Cost & $14.77$ & $14.28$ & $\mathbf{11.80}$ \\
 & Violation & $0.207$ & $0.201$ & $\mathbf{0.164}$ \\
 & Excess cost & $5.84$ & $5.72$ & $\mathbf{3.99}$ \\
\addlinespace
HalfCheetah & Return & $206.7$ & $\mathbf{220.9}$ & $218.5$ \\
 & Cost & $3.70$ & $3.89$ & $\mathbf{3.59}$ \\
 & Violation & $\mathbf{0.123}$ & $0.182$ & $0.158$ \\
 & Excess cost & $2.46$ & $2.20$ & $\mathbf{1.99}$ \\
\addlinespace
Ant & Return & $\mathbf{350.9}$ & $289.0$ & $286.4$ \\
 & Cost & $5.55$ & $\mathbf{3.54}$ & $4.08$ \\
 & Violation & $0.356$ & $0.210$ & $\mathbf{0.184}$ \\
 & Excess cost & $2.17$ & $\mathbf{1.56}$ & $2.23$ \\
\bottomrule
\end{tabular}
\end{adjustbox}
\end{table*}

\begin{table}[t]
\centering
\caption{A-Lag with its adaptive multiplier against A-Lag with a fixed multiplier, on the test tasks, with the same trained weights. The fixed value, given next to each environment, was chosen on training tasks by a fixed rule: the value with the lowest violation rate whose return is not clearly below that at $\lambda=0$ (Appendix~\ref{apx:controller}). Both columns are means over three seeds. Bold marks the better of the two in each row: the higher return and the lower cost, violation rate and excess cost. In CarGoal no tested value met the rule, so we use the one whose return came closest, $\lambda=0.3$.}
\label{tab:fixed}
\footnotesize
\setlength{\tabcolsep}{3pt}
\begin{tabular}{@{}llcc@{}}
\toprule
Environment & Metric & A-Lag, fixed $\lambda$ & A-Lag \\
\midrule
DarkRoom ($\lambda=0.3$) & Return & $0.831$ & $\mathbf{0.832}$ \\
 & Cost & $3.47$ & $\mathbf{3.45}$ \\
 & Violation & $0.190$ & $\mathbf{0.180}$ \\
 & Excess cost & $0.46$ & $\mathbf{0.39}$ \\
\addlinespace
PointGoal ($\lambda=1$) & Return & $0.150$ & $\mathbf{0.164}$ \\
 & Cost & $10.17$ & $\mathbf{9.69}$ \\
 & Violation & $0.137$ & $0.137$ \\
 & Excess cost & $4.43$ & $\mathbf{3.88}$ \\
\addlinespace
CarGoal ($\lambda=0.3$) & Return & $\mathbf{0.484}$ & $0.461$ \\
 & Cost & $13.29$ & $\mathbf{11.80}$ \\
 & Violation & $0.187$ & $\mathbf{0.164}$ \\
 & Excess cost & $5.06$ & $\mathbf{3.99}$ \\
\addlinespace
HalfCheetah ($\lambda=0.3$) & Return & $\mathbf{220.8}$ & $218.5$ \\
 & Cost & $3.94$ & $\mathbf{3.59}$ \\
 & Violation & $0.180$ & $\mathbf{0.158}$ \\
 & Excess cost & $2.25$ & $\mathbf{1.99}$ \\
\addlinespace
Ant ($\lambda=10$) & Return & $274.4$ & $\mathbf{286.4}$ \\
 & Cost & $\mathbf{3.49}$ & $4.08$ \\
 & Violation & $\mathbf{0.174}$ & $0.184$ \\
 & Excess cost & $\mathbf{1.78}$ & $2.23$ \\
\bottomrule
\end{tabular}
\end{table}

\section{Per-Seed Results}\label{apx:seeds}
Tables~\ref{tab:seed_alag} and~\ref{tab:seed_rq1} give the main comparisons for each training seed. Seed $s$ of one model is paired with seed $s$ of the other, on the same deployment draws. To describe uncertainty over test tasks, we resample the test tasks (10{,}000 bootstrap samples) and check whether the 95\% interval of the seed-averaged difference excludes zero. These intervals reflect which test tasks were drawn. Variation from retraining appears in the per-seed columns.

A-Lag's cost is lower than Safe AD's in all three seeds in CarGoal and Ant and in two of three in the other environments. The cost interval excludes zero in DarkRoom, CarGoal and Ant and includes it in PointGoal and HalfCheetah. Compared with $\lambda=0$, the multiplier lowers cost in all three seeds in DarkRoom, CarGoal and HalfCheetah, where the intervals exclude zero. Its violation-rate interval excludes zero in every environment but PointGoal. In Ant its cost interval lies above zero.

With dynamic budgets, the mean return is higher for both methods in every environment except A-Lag in Ant. It is higher in all three seeds for both methods in DarkRoom, PointGoal and CarGoal, and its interval excludes zero there for both methods and also in HalfCheetah and Ant for Safe AD. The cost is lower for both methods in DarkRoom and HalfCheetah, where the cost intervals exclude zero.

\begin{table}[t]
\centering
\caption{Per-seed differences on the test tasks: A-Lag against Safe AD, and A-Lag against its own weights with the multiplier held at zero. Each entry is the first model minus the second, averaged over the test tasks and deployments of one training seed.}
\label{tab:seed_alag}
\footnotesize
\setlength{\tabcolsep}{3pt}
\begin{tabular}{@{}llcccccc@{}}
\toprule
 & & \multicolumn{3}{c}{A-Lag $-$ Safe AD} & \multicolumn{3}{c}{A-Lag $-$ A-Lag, $\lambda=0$} \\
\cmidrule(lr){3-5}\cmidrule(lr){6-8}
Environment & Metric & Seed 1 & Seed 2 & Seed 3 & Seed 1 & Seed 2 & Seed 3 \\
\midrule
DarkRoom & Return & $-0.057$ & $-0.047$ & $-0.080$ & $-0.059$ & $-0.021$ & $-0.053$ \\
 & Cost & $-0.57$ & $+0.09$ & $-0.73$ & $-0.17$ & $-0.16$ & $-0.24$ \\
 & Violation & $-0.064$ & $-0.011$ & $-0.073$ & $-0.025$ & $-0.026$ & $-0.036$ \\
\addlinespace
PointGoal & Return & $+0.021$ & $-0.027$ & $-0.018$ & $-0.008$ & $+0.032$ & $+0.001$ \\
 & Cost & $-0.12$ & $-1.79$ & $+0.87$ & $-1.16$ & $-0.63$ & $+0.27$ \\
 & Violation & $-0.008$ & $-0.027$ & $+0.027$ & $-0.015$ & $-0.013$ & $+0.020$ \\
\addlinespace
CarGoal & Return & $-0.016$ & $-0.073$ & $+0.003$ & $-0.052$ & $-0.048$ & $-0.036$ \\
 & Cost & $-2.60$ & $-3.79$ & $-2.52$ & $-1.23$ & $-3.35$ & $-2.83$ \\
 & Violation & $-0.044$ & $-0.055$ & $-0.030$ & $-0.021$ & $-0.048$ & $-0.041$ \\
\addlinespace
HalfCheetah & Return & $-28.1$ & $+43.1$ & $+20.4$ & $-0.4$ & $-3.3$ & $-3.7$ \\
 & Cost & $+0.34$ & $-0.40$ & $-0.28$ & $-0.06$ & $-0.21$ & $-0.62$ \\
 & Violation & $0.000$ & $+0.015$ & $+0.090$ & $-0.006$ & $-0.018$ & $-0.048$ \\
\addlinespace
Ant & Return & $+14.4$ & $-76.4$ & $-131.4$ & $-1.2$ & $-2.0$ & $-4.6$ \\
 & Cost & $-0.65$ & $-1.10$ & $-2.68$ & $-0.06$ & $+0.22$ & $+1.45$ \\
 & Violation & $-0.051$ & $-0.086$ & $-0.380$ & $-0.052$ & $+0.001$ & $-0.027$ \\
\bottomrule
\end{tabular}
\end{table}

\begin{table}[t]
\centering
\caption{Per-seed differences on the test tasks between models trained on dynamic-budget and on fixed-budget histories. Each entry is the first model minus the second, averaged over the test tasks and deployments of one training seed.}
\label{tab:seed_rq1}
\footnotesize
\setlength{\tabcolsep}{3pt}
\begin{tabular}{@{}llcccccc@{}}
\toprule
 & & \multicolumn{3}{c}{Safe AD, dynamic $-$ fixed} & \multicolumn{3}{c}{A-Lag, dynamic $-$ fixed} \\
\cmidrule(lr){3-5}\cmidrule(lr){6-8}
Environment & Metric & Seed 1 & Seed 2 & Seed 3 & Seed 1 & Seed 2 & Seed 3 \\
\midrule
DarkRoom & Return & $+0.043$ & $+0.147$ & $+0.311$ & $+0.085$ & $+0.134$ & $+0.152$ \\
 & Cost & $-0.78$ & $-0.91$ & $-0.74$ & $-1.19$ & $+0.16$ & $-1.49$ \\
 & Violation & $-0.080$ & $-0.104$ & $-0.071$ & $-0.128$ & $-0.012$ & $-0.152$ \\
\addlinespace
PointGoal & Return & $+0.081$ & $+0.078$ & $+0.059$ & $+0.105$ & $+0.093$ & $+0.018$ \\
 & Cost & $+0.53$ & $+0.77$ & $+1.23$ & $+2.08$ & $+2.12$ & $+0.97$ \\
 & Violation & $+0.017$ & $+0.012$ & $+0.009$ & $+0.028$ & $+0.029$ & $+0.026$ \\
\addlinespace
CarGoal & Return & $+0.224$ & $+0.279$ & $+0.238$ & $+0.227$ & $+0.168$ & $+0.218$ \\
 & Cost & $+1.51$ & $+0.80$ & $+2.31$ & $-1.36$ & $-3.35$ & $-0.80$ \\
 & Violation & $+0.025$ & $+0.010$ & $+0.035$ & $-0.018$ & $-0.033$ & $-0.003$ \\
\addlinespace
HalfCheetah & Return & $+28.7$ & $-11.0$ & $+92.3$ & $-10.5$ & $-63.4$ & $+97.9$ \\
 & Cost & $-0.39$ & $-0.58$ & $-1.47$ & $-2.52$ & $-1.84$ & $+2.04$ \\
 & Violation & $-0.007$ & $-0.028$ & $-0.024$ & $-0.038$ & $-0.098$ & $+0.116$ \\
\addlinespace
Ant & Return & $+36.9$ & $-1.8$ & $+80.2$ & $+20.9$ & $+18.1$ & $-79.6$ \\
 & Cost & $+1.29$ & $-1.95$ & $+3.67$ & $-2.19$ & $+0.15$ & $-1.03$ \\
 & Violation & $+0.084$ & $-0.135$ & $+0.300$ & $-0.115$ & $+0.031$ & $-0.127$ \\
\bottomrule
\end{tabular}
\end{table}

\section{Safe AD on PPO-Lagrangian Learning Histories}\label{apx:ppolag}
\begin{figure*}[t]
  \centering
  \includegraphics[width=\textwidth]{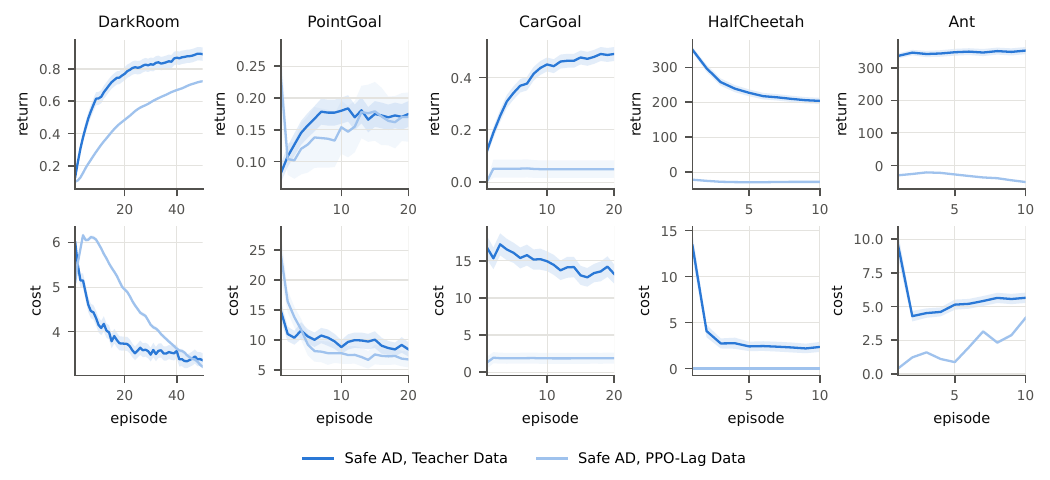}
  \caption{Safe AD trained on the teacher data (Teacher Data)
  and the Safe AD baseline of \citet{moeini2026safeincontextreinforcementlearning}, trained on PPO-Lagrangian learning histories (PPO-Lag
  Data). }
  \label{fig:ppolag}
\end{figure*}

Section~\ref{sec:data} notes that the learning histories of a safe RL
collector are not built to show the behavior a Safe ICRL agent needs.
Figure~\ref{fig:ppolag} compares the main Safe AD, trained on the teacher
data, with the Safe AD baseline of \citet{moeini2026safeincontextreinforcementlearning}, trained on learning histories of
PPO-Lagrangian~\citep{ray2019benchmarking}.
In DarkRoom, HalfCheetah and Ant its curves are per-episode means from its
original evaluation. In PointGoal and CarGoal they come from a later
evaluation of the same models over 20 episodes. Both evaluations used their
own test tasks and targets rather than the protocol of
Section~\ref{sec:experiments}. The comparison is therefore not paired, and its
differences cannot be attributed to the source of the histories. It was also trained on 8 tasks per environment with older
training settings.

In DarkRoom the earlier model learns the task, but its return stays below that
of the main Safe AD in every episode, and its cost stays higher until the last
episodes. In HalfCheetah and Ant its return is negative in every episode, far
below the main Safe AD, and it never pays any cost in HalfCheetah. Its PointGoal
return is close to that of the main Safe AD, with a slightly lower final cost,
while its CarGoal return stays near 0.05, far below the main Safe AD, at a far
lower cost.

\end{document}